\documentclass[runningheads]{llncs}
\usepackage[T1]{fontenc}
\usepackage{graphicx}

\usepackage[outdir=./]{epstopdf}

\usepackage{setspace}
\usepackage{subfiles}
\usepackage{multirow} %
\usepackage{rotating} %
\usepackage{lscape}
\usepackage{makecell}
\usepackage{tikz}
\usepackage{float}
\usepackage{pifont}
\usepackage{placeins}
\usepackage{bibnames}

\usepackage{booktabs} %
\usepackage{microtype} %
\usepackage[colorlinks]{hyperref}
\usepackage[export]{adjustbox}
\usepackage{array}
\usepackage{rotating}

\usepackage{color}

\begin{document}
\title{Sanity Checks for Saliency Methods Explaining Object Detectors}

\author{Deepan Chakravarthi Padmanabhan\inst{1}\orcidID{0000-0003-0638-014X} \and
Paul G. Pl{\"o}ger\inst{1}\orcidID{0000-0001-5563-5458} \and
Octavio Arriaga\inst{2}\orcidID{0000-0002-8099-2534} \and
Matias Valdenegro-Toro\inst{3}\orcidID{0000-0001-5793-9498}}
\authorrunning{DC. Padmanabhan et al.}

\institute{Bonn-Rhein-Sieg University of Applied Sciences, Sankt Augustin, Germany \email{deepangrad@gmail.com}
 \and University of Bremen, Bremen, Germany \email{arriagac@uni-bremen.de} \and University of Groningen, Groningen, The Netherlands \email{m.a.valdenegro.toro@rug.nl}}

\maketitle %

\begin{abstract}
Saliency methods are frequently used to explain Deep Neural Network-based models. Adebayo \textit{et al.}'s work on evaluating saliency methods for classification models illustrate certain explanation methods fail the model and data randomization tests. However, on extending the tests for various state of the art object detectors we illustrate that the ability to explain a model is more dependent on the model itself than the explanation method. We perform sanity checks for object detection and define new qualitative criteria to evaluate the saliency explanations, both for object classification and bounding box decisions, using Guided Backpropagation, Integrated Gradients, and their Smoothgrad versions, together with Faster R-CNN, SSD, and EfficientDet-D0, trained on COCO. In addition, the sensitivity of the explanation method to model parameters and data labels varies class-wise motivating to perform the sanity checks for each class. We find that EfficientDet-D0 is the most interpretable method independent of the saliency method, which passes the sanity checks with little problems.

\end{abstract}

\section{Introduction}

Localizing and categorizing different object instances is pivotal in various real-world applications such as autonomous driving \cite{Feng_ADsurvey}, healthcare \cite{Araujo_UOLO}, and text detection \cite{He_TextDetector}.
Recent advances with Deep Neural Network-based (DNN) object detectors demonstrate remarkable performances both in terms of robustness and generalization across practical use cases \cite{Arani_ODSurvey}.
Even though detectors are extensively needed in safety-critical applications, the heavily parameterized DNN-based detectors limit understanding the rationale behind the detections made by such detectors.
In addition, object detectors are prone to non-local effects as a slight change in the object position can affect the detector prediction \cite{Rosenfeld_Elephant}.
Therefore, explaining detector decisions is imperative to earn user trust and understand the reason behind predictions to a certain extent in safety-critical situations, overall improving system safety.

Explaining a DNN decision-making process has been addressed prominently \cite{Simonyan_Gradients} \cite{Zeiler_DeconvNet} \cite{Springenberg_GuidedBackpropagation} \cite{Bach_LRP}. 
The explanations are useful for debugging the model, reveal the spurious effects and biases learned by a model as well as underpins regulatory requirements (like GDPR). 
Furthermore, such explanations boost transparency and contribute towards safety of the associated DNN-based systems \cite{Huang_aisafety} \cite{Toner_aisafety}.
Among the methods explaining DNNs, saliency methods are popular explanation methods \cite{Samek_CVPRW21} \cite{Marcinkevics_IMLZoo}, which provide the input feature attribution that highlights the most relevant pixels responsible for the model prediction. 
Despite extensive study of employing saliency methods to classification tasks, only handful of works explain detector decisions \cite{Petsiuk_DRISE} \cite{Tsunakawa_SSD_CRP} \cite{Gudovskiy_E2X}. 
Moreover, the evaluation metrics used to quantitatively assess the detector explanations fail certain sanity checks as well and prove to be statistically unreliable \cite{Tomsett_sanity_checks_for_saliency_metrics}.

Sanity checks are basic procedures to test the ability of an explanation method to correctly explain a model decision \cite{Adebayo_sanitychecks} or test the ability of an evaluation metrics to correctly assess the explanation method \cite{Tomsett_sanity_checks_for_saliency_metrics} that generates a saliency map.
In this paper, we are concerned with the former, where we check the ability of an explanation method to generate relevant saliency map based explanations for detections made by an object detector. 
However, there is limited work studying object detector explainability, and in particular basic sanity checks have not been performed to the best of our knowledge.
Therefore, conducting simple sanity checks to determine the quality of an explanation method is extremely important.

In this paper we conduct simple sanity checks for certain explanation methods explaining three object detector predictions.
We extend the sanity checks in \cite{Adebayo_sanitychecks} to object detectors.
The sanity checks test explanation method sensitivity towards the detectors parameters (model randomization test) and data generation method (data randomization test). 

\begin{figure}[!t]
	\centering
	\includegraphics[width=\linewidth]{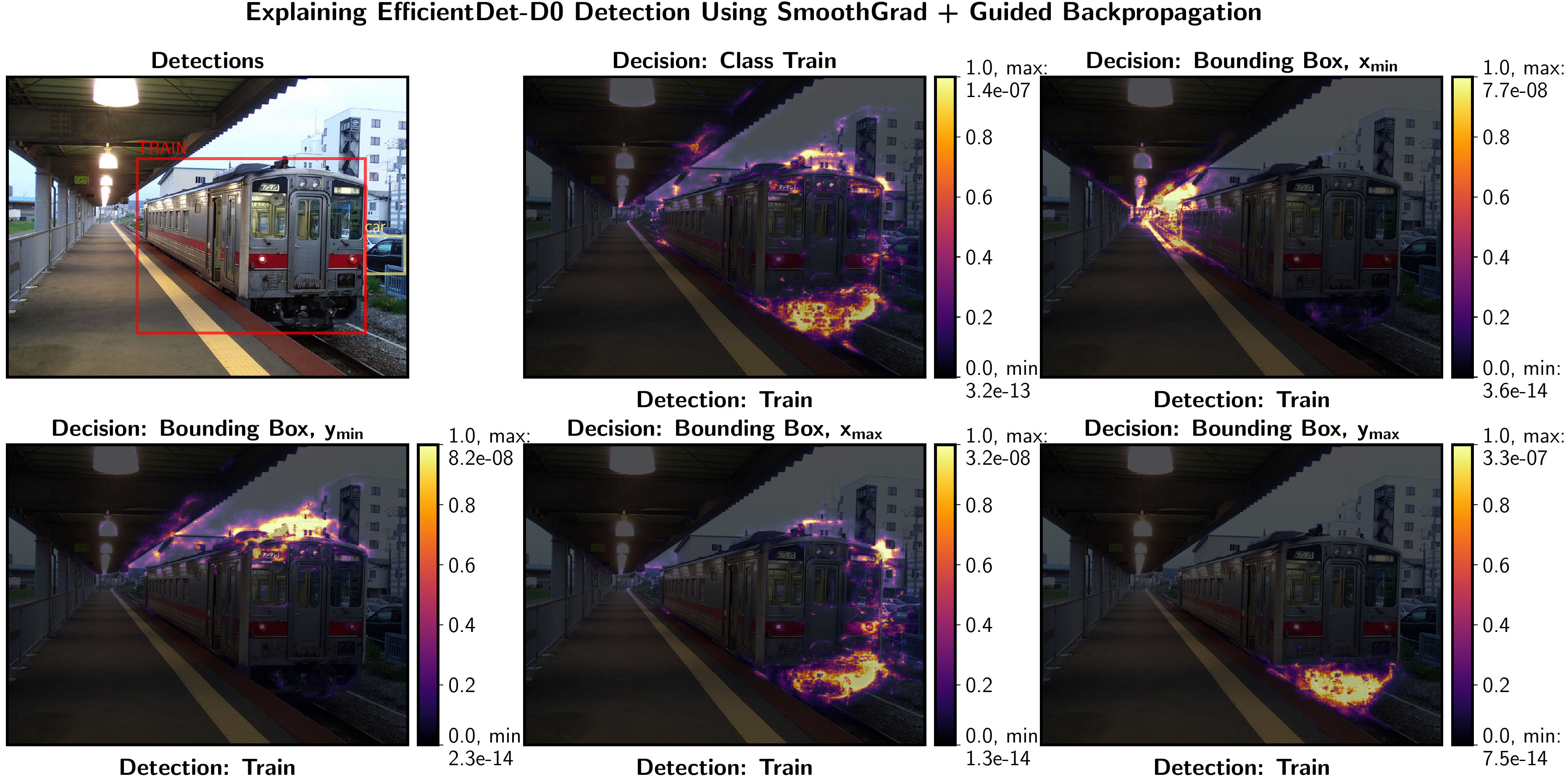}
	\caption{Sample detection explanations using EfficientDet-D0 and SGBP, considering one saliency explanation for classification and bounding box regression decisions. We find that EfficientDet-D0 provides high quality explanations that pass sanity checks. For all figures in this paper, the saliency maps are overlaid on the corresponding original image after min-max normalization with the minimum and maximum value indicated in the corresponding heatmap.}
	\label{fig:explanation_sample}
\end{figure}
The contributions of our paper are:
\begin{itemize}
	\item We evaluate sanity checks for saliency explanations of object detectors, both on classification and bounding box decision explanations.
	\item We define clear qualitative evaluation criteria for sanity checks in saliency explanations for object detectors.
	\item We find that Modern object detectors like EfficientDet-D0 \cite{Tan_EfficientDet} seem to be more interpretable and pass more sanity checks than older detectors like SSD \cite{Liu_SSD} and Faster R-CNN \cite{Ren_FasterRCNN}.
\end{itemize}
We expect that our work helps advance our understanding of object detector explainability and increases the use of explanations in computer vision.

\section{Related Work}
Adebayo \textit{et al.} \cite{Adebayo_sanitychecks} are the pioneers to propose sanity checks for explanation methods based on randomization tests. 
The authors identify that various widely used explanation methods provide saliency map explanations that are independent of the model parameters and the data used to develop the model.
The widely-used gradient-based explanation methods such as guided backpropagation \cite{Springenberg_GuidedBackpropagation} and Guided GradCAM \cite{Selvaraju_GradCAM} fail both model and data randomization sanity checks. 
In this related work, the sanity checks are performed on classifier models such as Inception v3, CNN, and MLP trained using ImageNet, Fashion MNIST, and MNIST datasets respectively. 
However, Yona \textit{et al.} \cite{Yona_revisiting_sanity_checks} posit that the randomization tests are distribution-dependent and modify the sanity checks proposed in \cite{Adebayo_sanitychecks} with a causal perspective. 
The model sensitivity test is performed by combining the original images with multiple or partial objects to generate saliency maps for random and trained model. 
This reformulation is an attempt to spatially control the relevant features for a particular class and extract visually distinct saliency maps. 
The methods failing the sanity checks in \cite{Adebayo_sanitychecks} such as vanilla and guided backpropagation pass this reformulated version.
Kindermans \textit{et al.} \cite{Kindermans_unreliability} proposes input invariance property as a sanity check for saliency methods. 
The saliency method output should not be affected by the transformations done to the input, mirroring the model sensitivity to the specific transformation. 
Experiments on MNIST illustrate the possibility to forcefully manipulate the explanations. 
The literature on interpretability cover certain axioms such as completeness \cite{Bach_LRP}, implementation invariance, and sensitivity \cite{Sundararajan_IG} are considered as indicators of reliability for saliency methods. 
Kim \textit{et al.} \cite{Kim_sanity_simulations} develop a synthetic benchmark and enable a ground-truth-based evaluation procedure. 
Various evaluation metrics to assess the explanation method with regards to factors such as faithfulness, robustness, and fairness of explanation is provided by \cite{Quantus}.
Tomsett \textit{et al.} \cite{Tomsett_sanity_checks_for_saliency_metrics} conclude the evaluation metrics assessing the faithfulness of the explanations are unreliable by conducting certain sanity checks on the metrics. 
In this paper, we extend the sanity checks performed by \cite{Adebayo_sanitychecks} based on randomization to detectors and report our findings.

\section{Sanity Checks for Object Detection Saliency Explanations}
\label{ex:sanity_checks}

We use two kinds of sanity checks as defined by Adebayo \textit{et al.} \cite{Adebayo_sanitychecks}. The model parameter and data randomization tests have been proposed to evaluate the explanation methods for classification tasks. 

\textbf{Model Randomization}. The model parameter randomization test analyzes the saliency method output for a trained classifier model against the saliency method output for a model parameter initialized with random values \cite{Adebayo_sanitychecks}. 
The saliency maps help to understand the explanation method sensitivity to model parameters and to model properties, in general. 
A similar saliency map signifies that the saliency method will not be helpful to debug a model as the saliency method is invariant to the model parameters.

\textbf{Data Randomization}. In the data randomization test, the saliency maps for a model trained on a correctly labeled dataset and model trained using randomly permuted labels are compared \cite{Adebayo_sanitychecks}.

A similar saliency map between the two outputs illustrates the relationship insensitivity between labels and input images. 
The saliency maps will not reflect the reason behind label and input image relationship captured by the data generation process.
If the explanations are indifferent to a random label assigned to a mammogram image, for instance, the saliency map fails to explain the real reason for a diagnosis output.

The tests serve as sanity checks to assess the scope of a particular explanation method for explaining models performing certain tasks. These are very basic assumptions made on saliency explanations and many methods fail these basic tests in classification tasks.

In this paper, we use the two randomization tests on pre-trained object detectors, for a certain set of saliency explanation methods, and we test if those detectors and explanation methods pass the basic sanity checks.

\begin{table*}[t]    
	\centering
	\caption[Subjective analysis of sanity check results]{Summary of the subjective analysis for the model randomization test.
		The score is computed as explained in Sec.~\ref{section:sanity_scores}.
		The higher the score the more sensitive is the method for the detector model parameters.
		Each column indicates an aspect considered to evaluate the change in the saliency map which is produced for the randomized model.
		The table is generated by scoring the majority characteristic illustrated by each detector and explanation method combination over 15 randomly sampled detections from the COCO test 2017 split.}
	\begin{tabular}{lllllllll} 
		\toprule
		\textbf{OD} & \textbf{IM} & \rotatebox{75}{\textbf{\shortstack[l]{Edge\\Detector}}} & \rotatebox{75}{\textbf{\shortstack[l]{ Highlight only\\Interest Object}}} & \rotatebox{75}{\textbf{\shortstack[l]{Focus more\\than One Object}}} & \rotatebox{75}{\textbf{\shortstack[l]{Texture\\Change }}} & \rotatebox{75}{\textbf{\shortstack[l]{Illustrate\\Artifacts}}}
		& \rotatebox{75}{\textbf{\shortstack[l]{Intensity\\range change}}} & \rotatebox{75}{\textbf{Score}}\\ 
		\midrule
		\multirow{4}{*}{ED0} & GBP & \ding{55} & \ding{55} & \ding{55} &\ding{51} & \ding{55} & \ding{51}& 7 \\
		& SGBP & \ding{55} & \ding{55} & \ding{55} & \ding{51} & \ding{55}& \ding{51}& 7 \\
		& IG & \ding{55} & \ding{55}& \ding{55} & \ding{51}& \ding{55} & \ding{51}& 7\\
		& SIG & \ding{55} & \ding{55}  &\ding{55} & \ding{51} & \ding{55} & \ding{51}& 7\\
		\midrule
		\multirow{4}{*}{SSD} & GBP & \ding{51} & \ding{55} &  \ding{55} & \ding{51} & \ding{55}& \ding{51} & 5\\
		& SGBP & \ding{51} & \ding{55} & \ding{55} & \ding{51} & \ding{55} & \ding{51}& 5\\
		& IG & \ding{51}& \ding{55} & \ding{55} & \ding{51}& \ding{55} & \ding{51}& 5\\
		& SIG & \ding{51} & \ding{55} &  \ding{55} & \ding{51} & \ding{55}& \ding{51}& 5 \\ 
		\midrule
		\multirow{4}{*}{FRN}  & GBP & \ding{55} & \ding{55} & \ding{55} & \ding{55} &\ding{51}& \ding{51} & 1\\
		& SGBP & \ding{55} & \ding{55} & \ding{55} & \ding{55} & \ding{51} & \ding{51}& 1\\
		& IG & \ding{55} & \ding{55} &  \ding{51} & \ding{51}&\ding{51}& \ding{51} & 5\\
		& SIG & \ding{55} &\ding{55}& \ding{51} & \ding{51} & \ding{51} & \ding{51}& 5\\
		\bottomrule
	\end{tabular}
	\label{tab:subjective_analysis}
\end{table*}

\begin{table*}[!htp]
    \centering
    \caption{Visual illustrations of saliency map sanity check properties. This table compares explanation patterns made by different detectors and saliency explanation methods against a randomly trained model. These results complement the qualitative evaluation we perform in this paper.}
    \begin{tabular}{llll}
        & Edge Detector & Highlight object of  & Focus certain \\            
        & (SSD\_SGBP) & interest (SSD\_GBP) & objects (FRN\_IG) \\
        
        \rotatebox[origin=c]{90}{Detection} &
        \includegraphics[scale=0.30,valign=c]{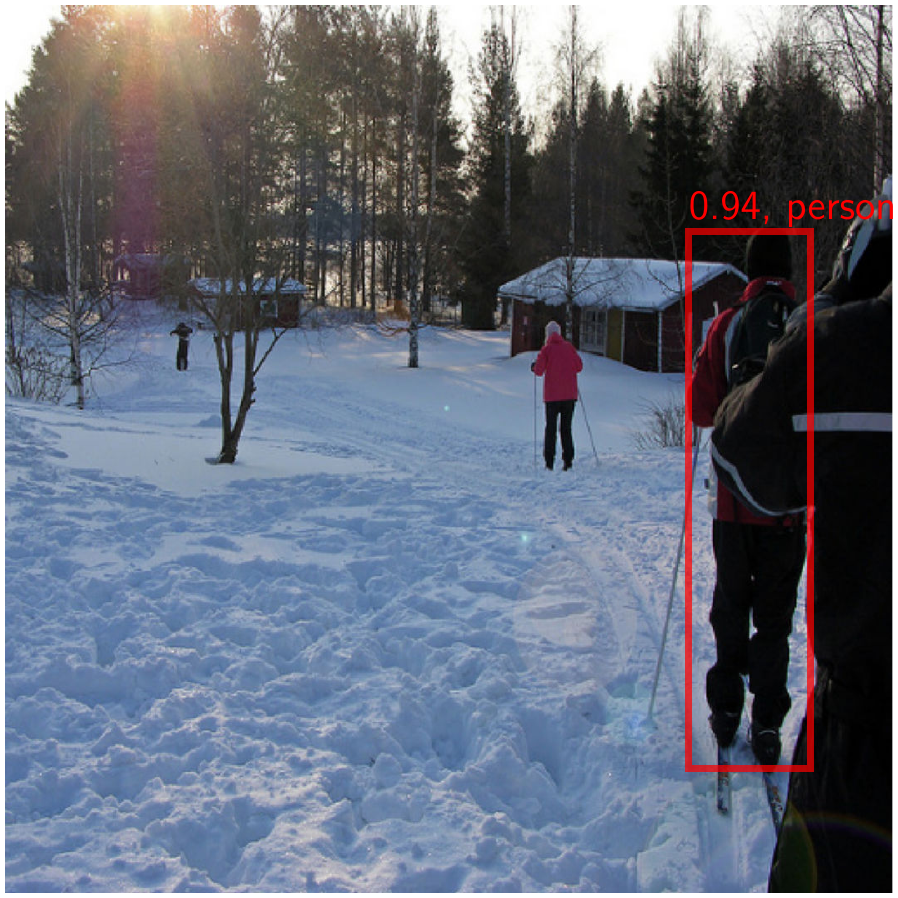}   &
        \includegraphics[scale=0.30,valign=c]{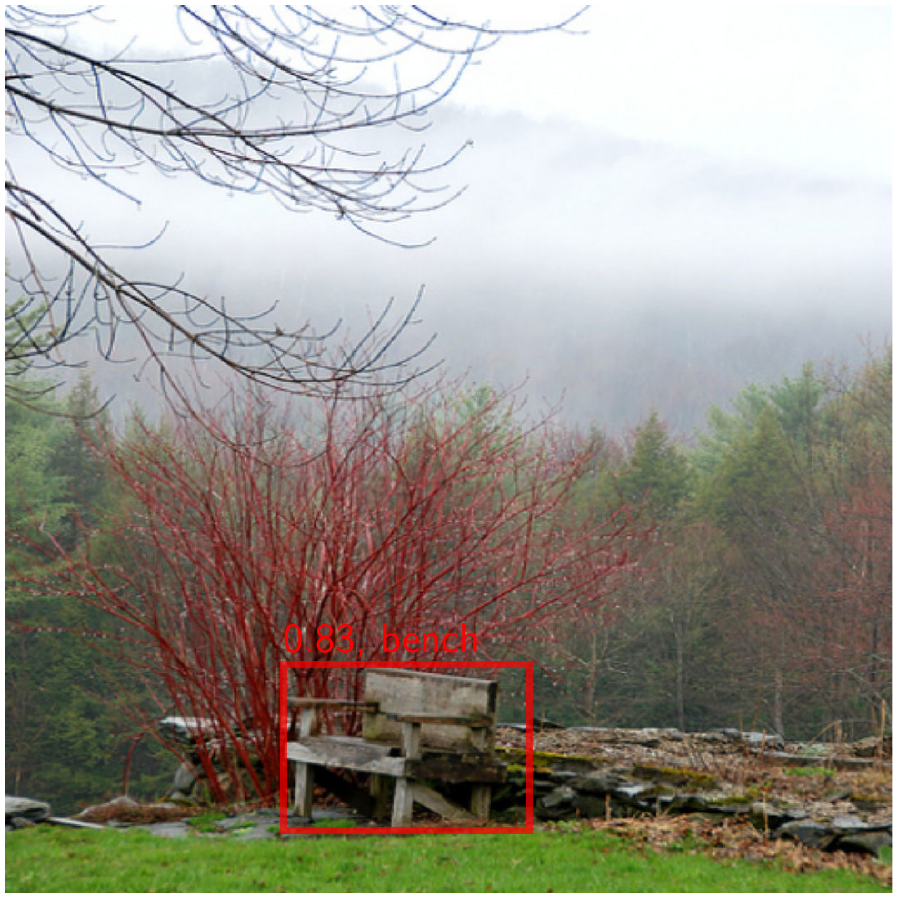}   &
        \includegraphics[scale=0.30,valign=c]{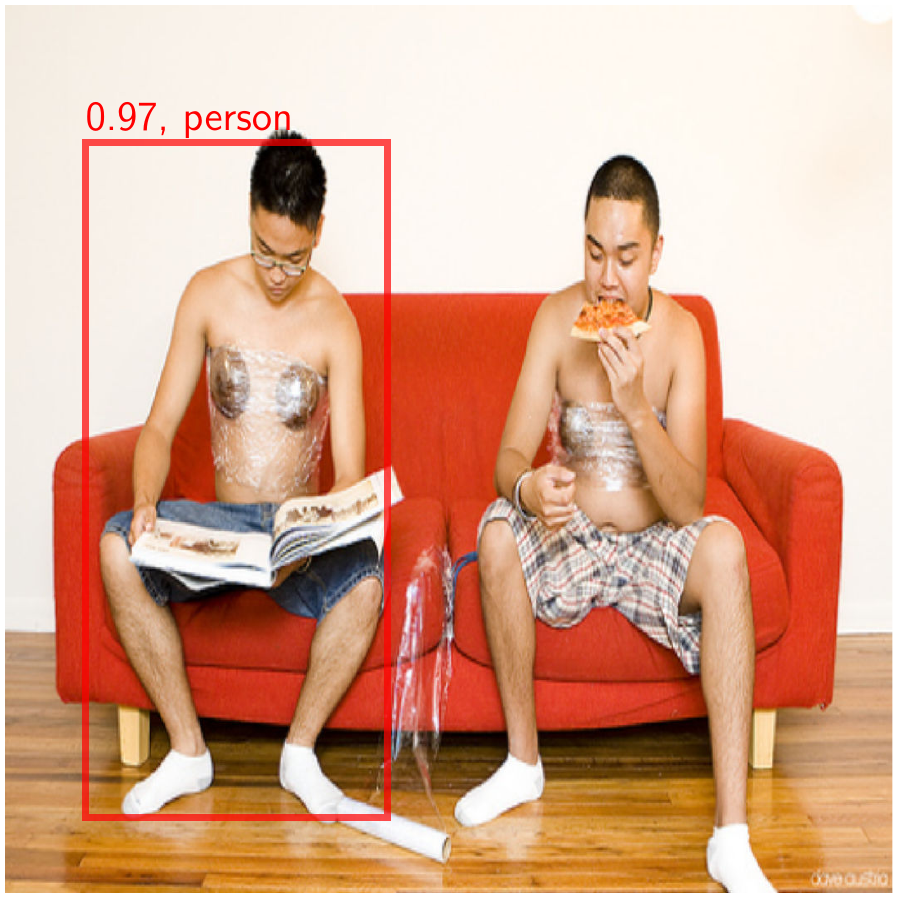} 			
        \\[1em]
        
        \rotatebox[origin=c]{90}{True model} &
        \includegraphics[scale=0.30,valign=c]{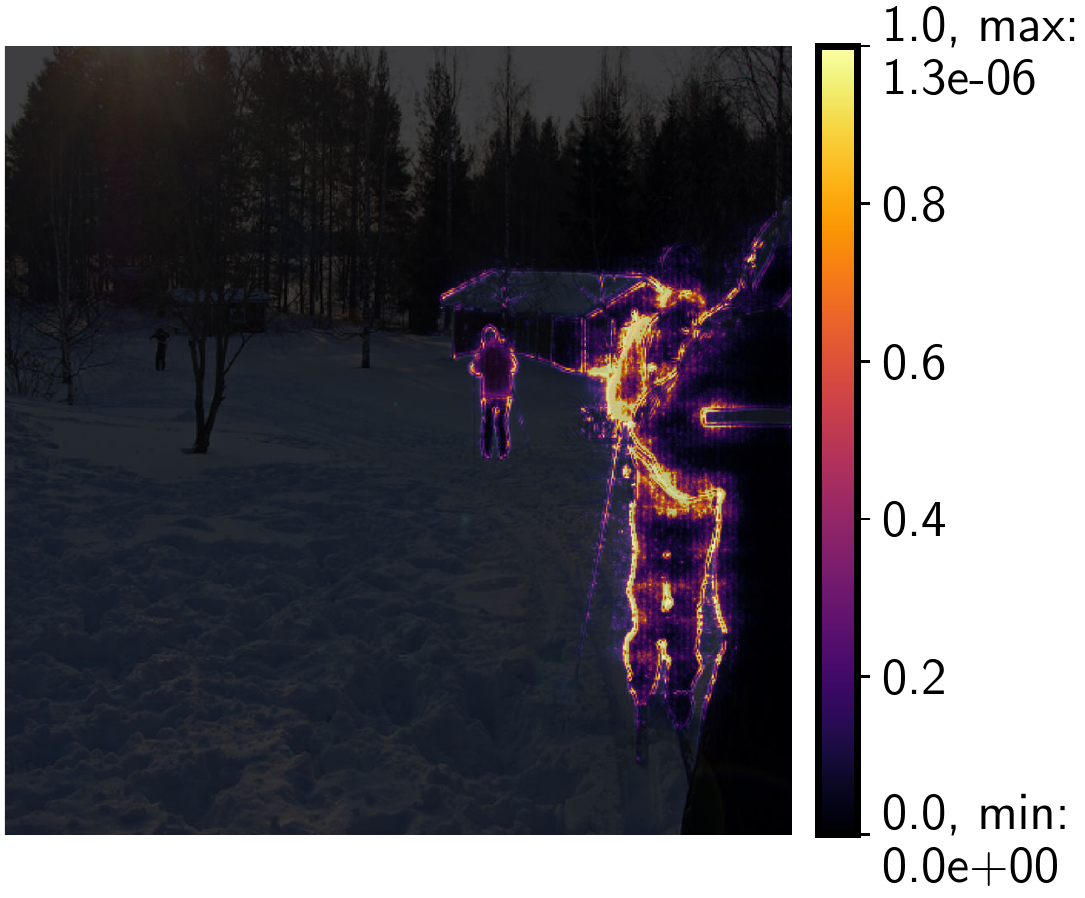}   &
        \includegraphics[scale=0.30,valign=c]{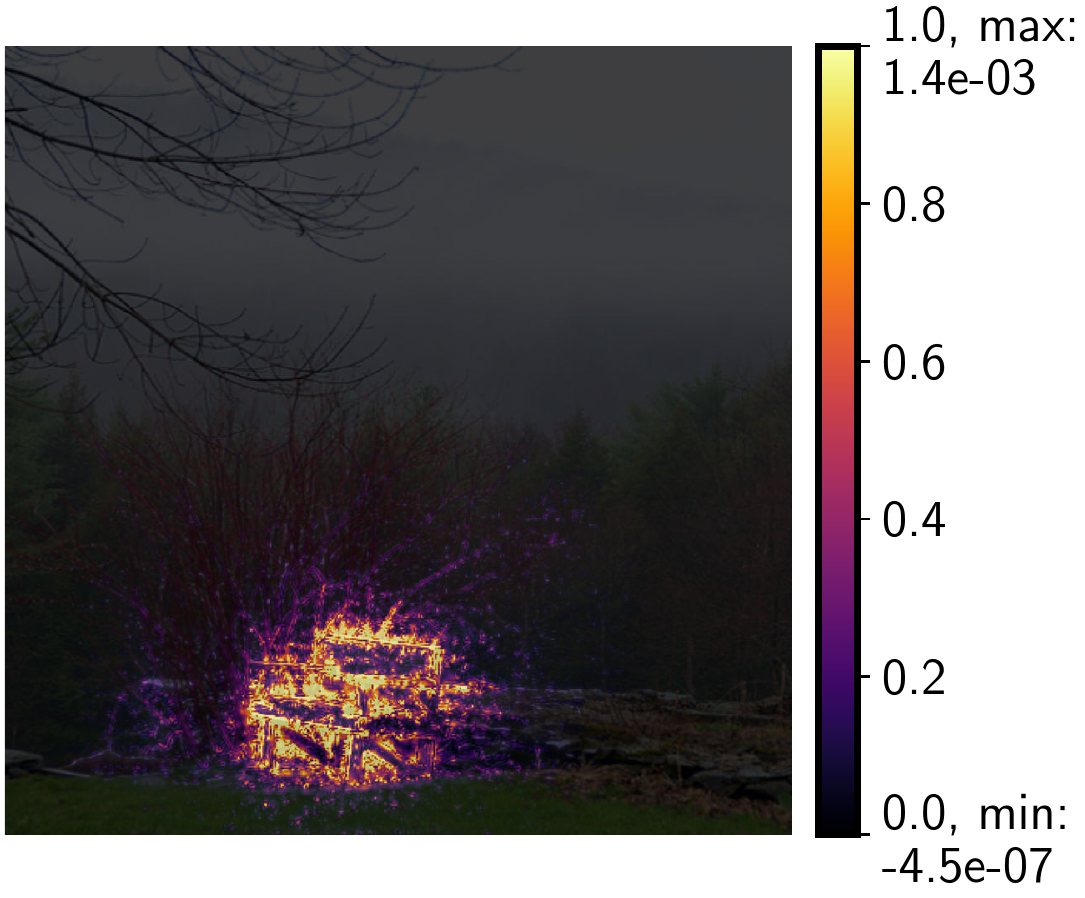}   &
        \includegraphics[scale=0.30,valign=c]{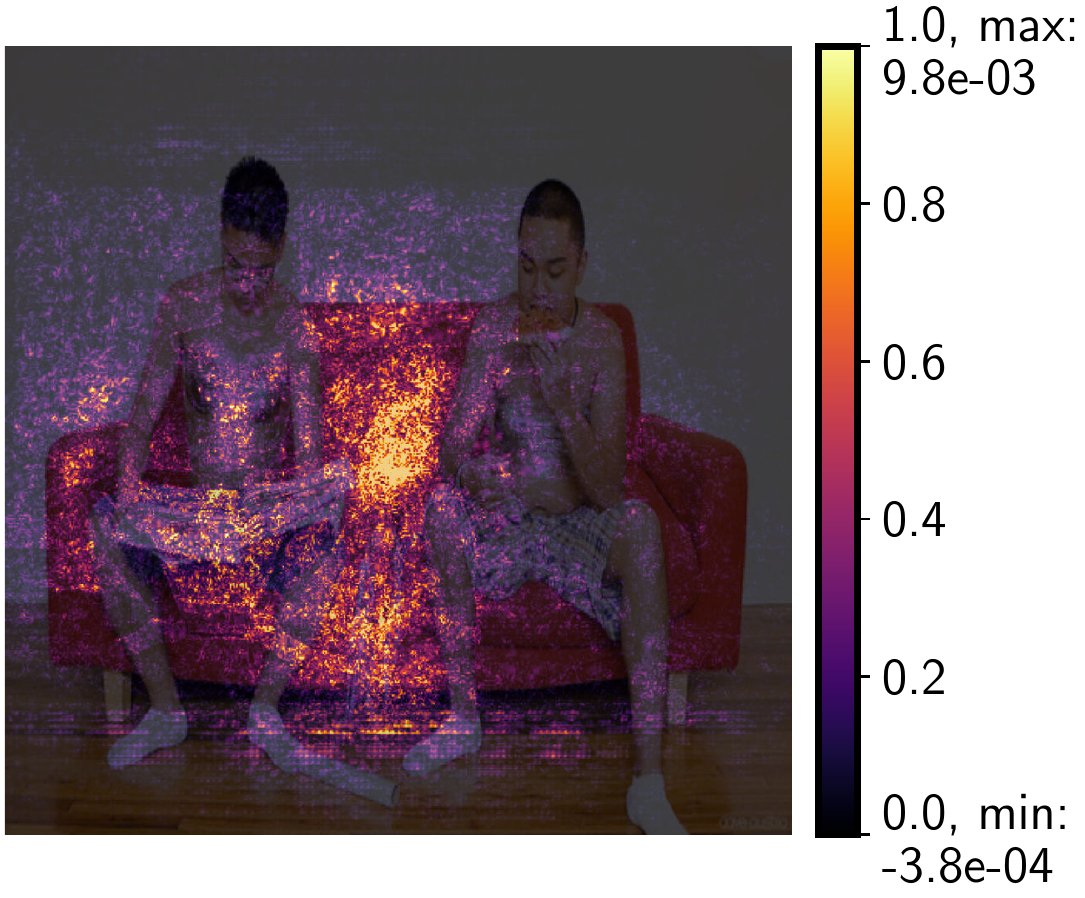}  		
        \\[1em]
        
        \rotatebox[origin=c]{90}{Random model} & 
        \includegraphics[scale=0.30,valign=c]{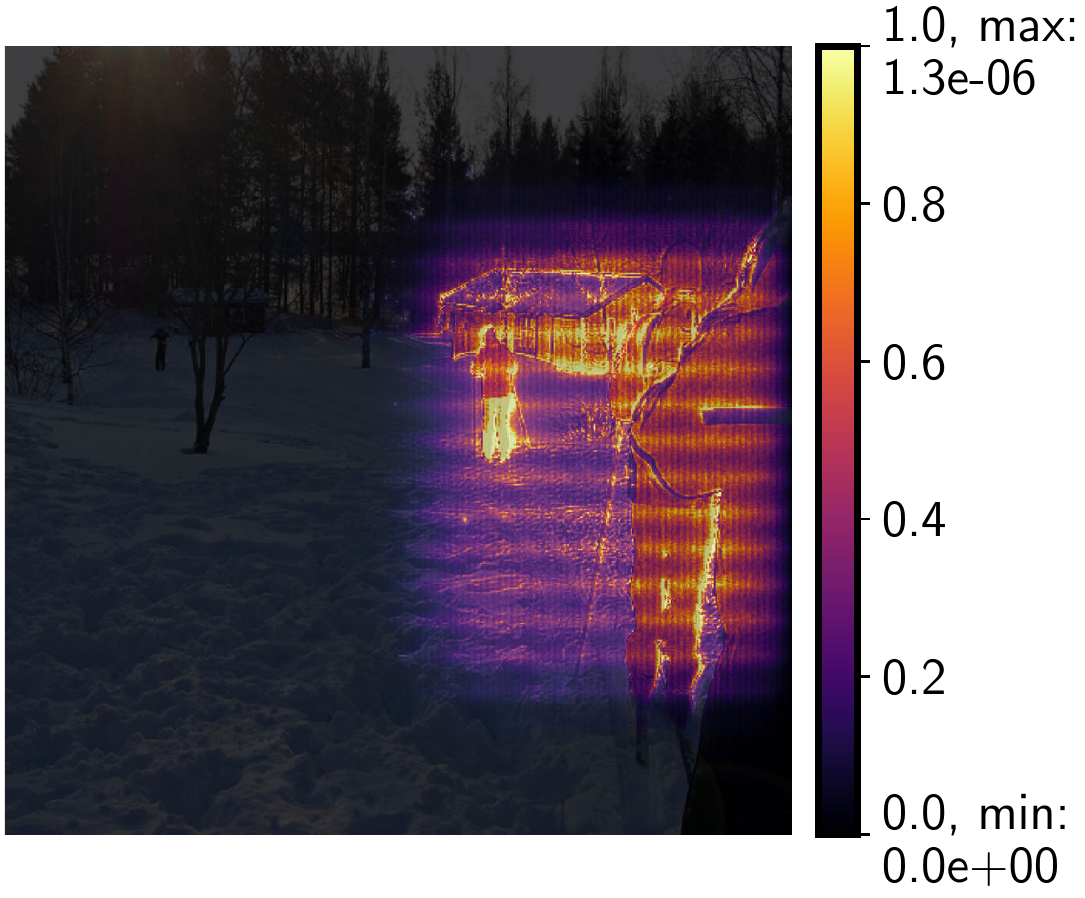}  &
        \includegraphics[scale=0.30,valign=c]{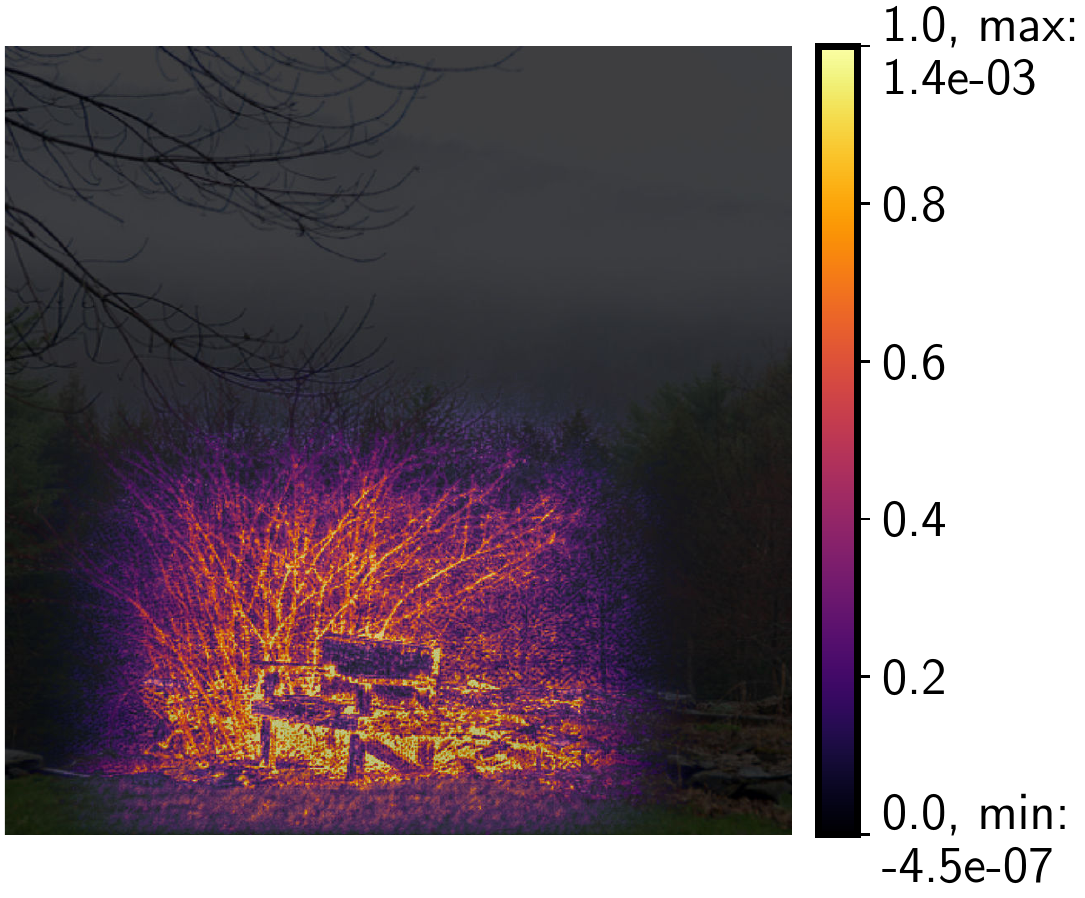}   &
        \includegraphics[scale=0.30,valign=c]{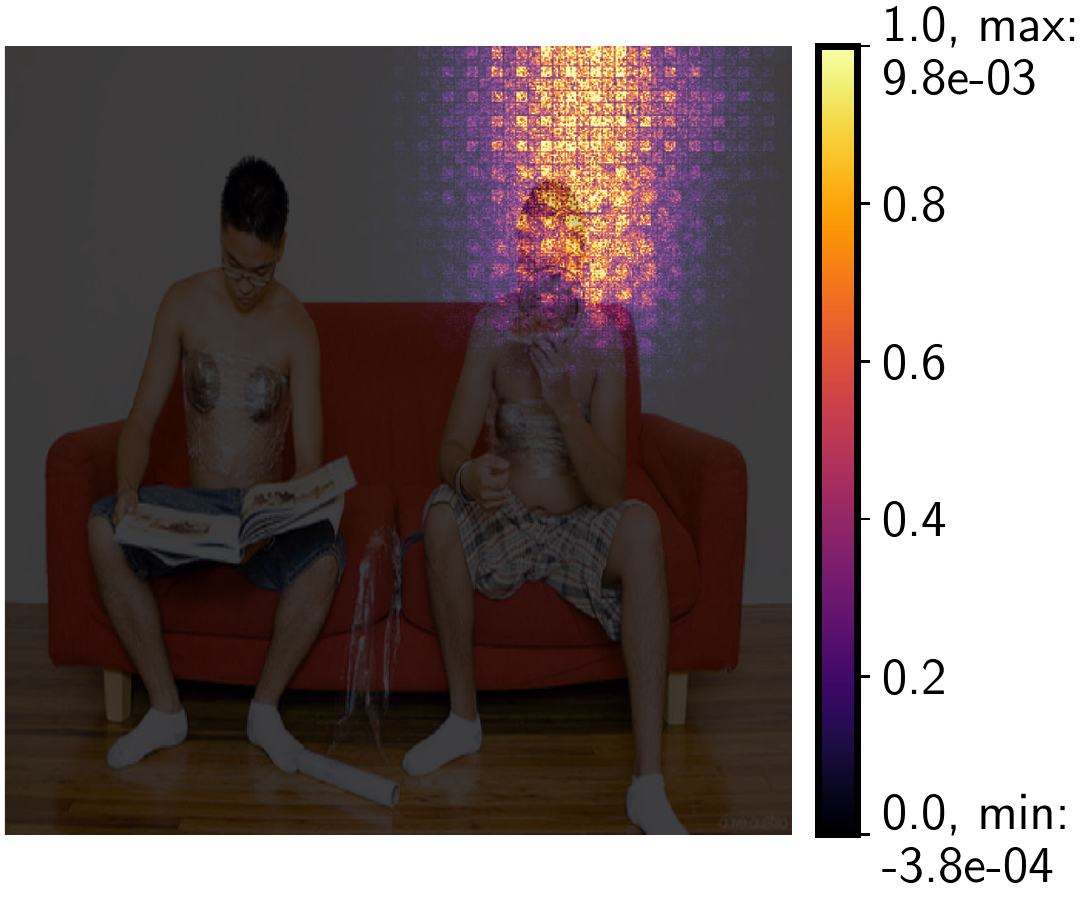}  \\
        
        & Texture change & Illustrate artifact & Intensity change \\
        & (ED0\_GBP) & (FRN\_SGBP) & (ED0\_SIG)\\
        
        \rotatebox[origin=c]{90}{Detection} &
        \includegraphics[scale=0.30,valign=c]{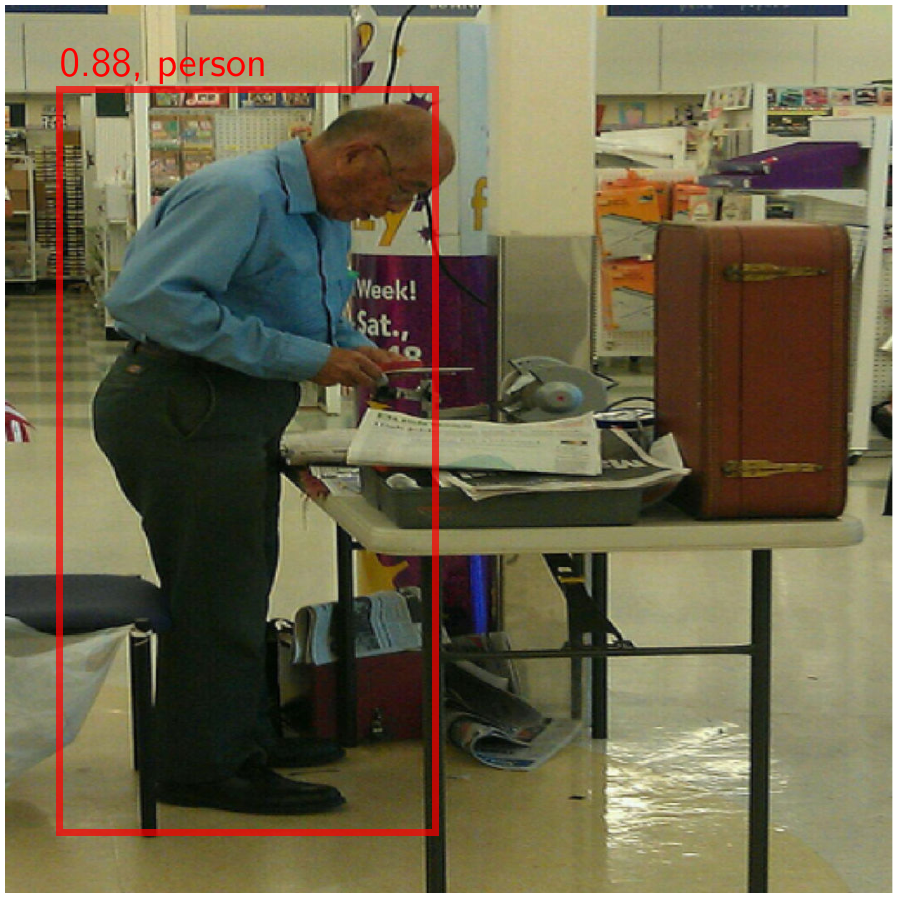}   &
        \includegraphics[scale=0.30,valign=c]{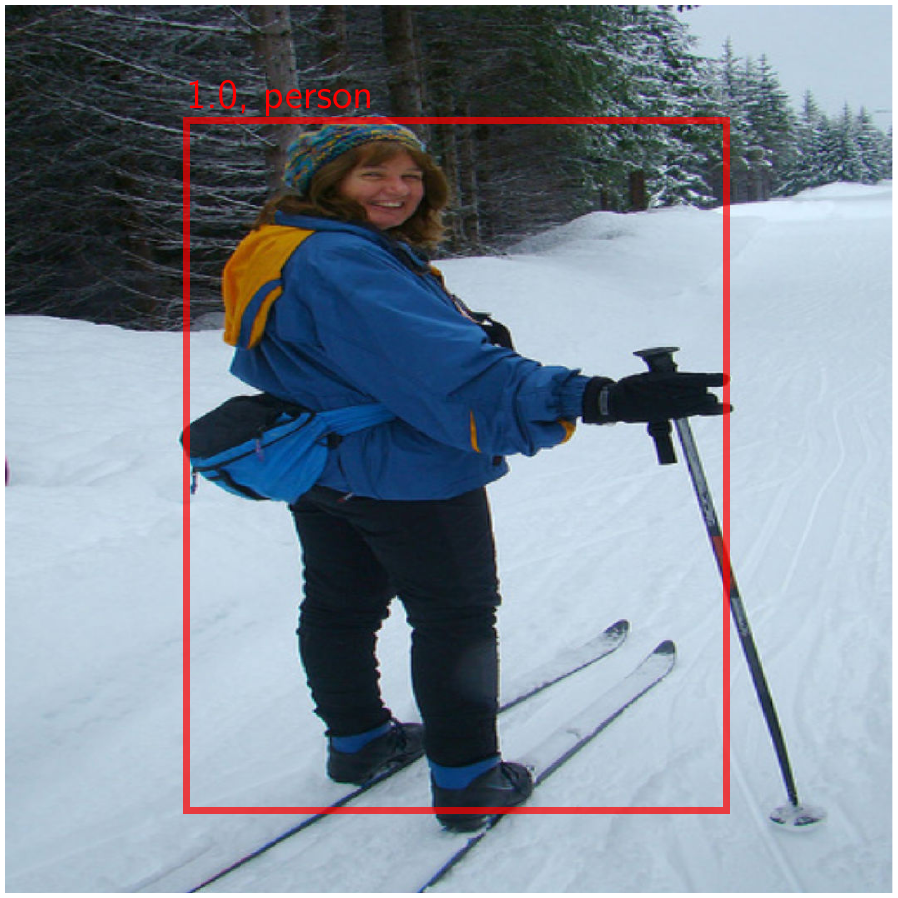}   &
        \includegraphics[scale=0.30,valign=c]{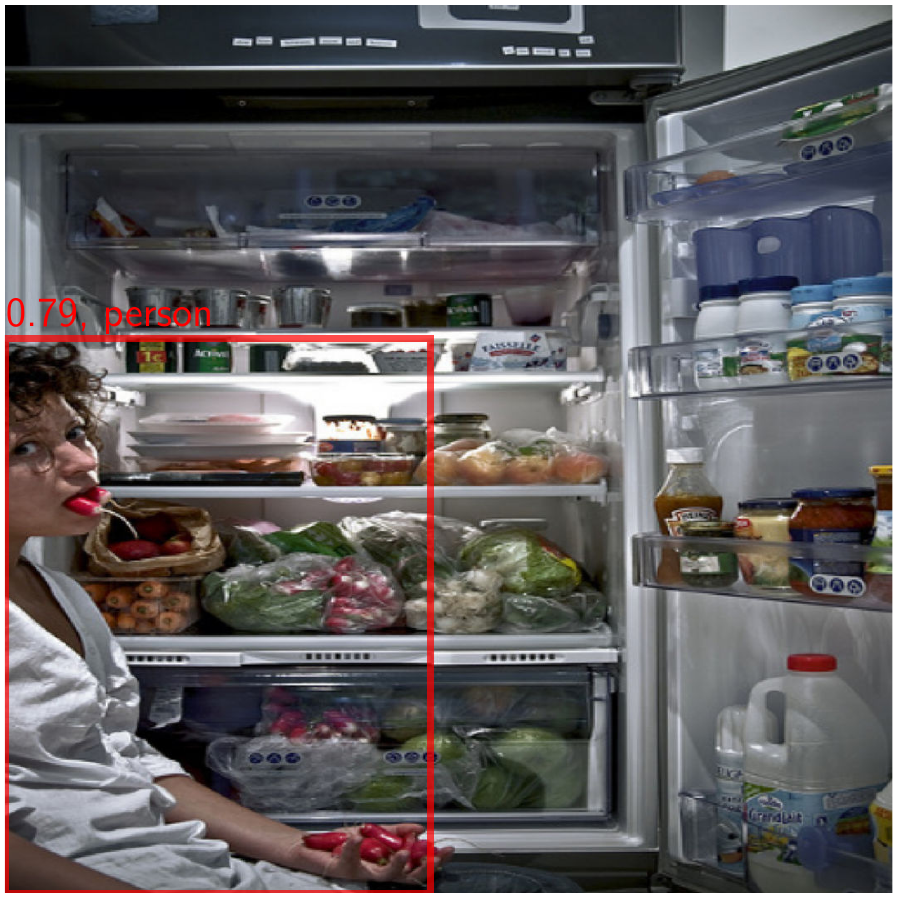}
        \\[1em]
        
        \rotatebox[origin=c]{90}{True model} &
        \includegraphics[scale=0.30,valign=c]{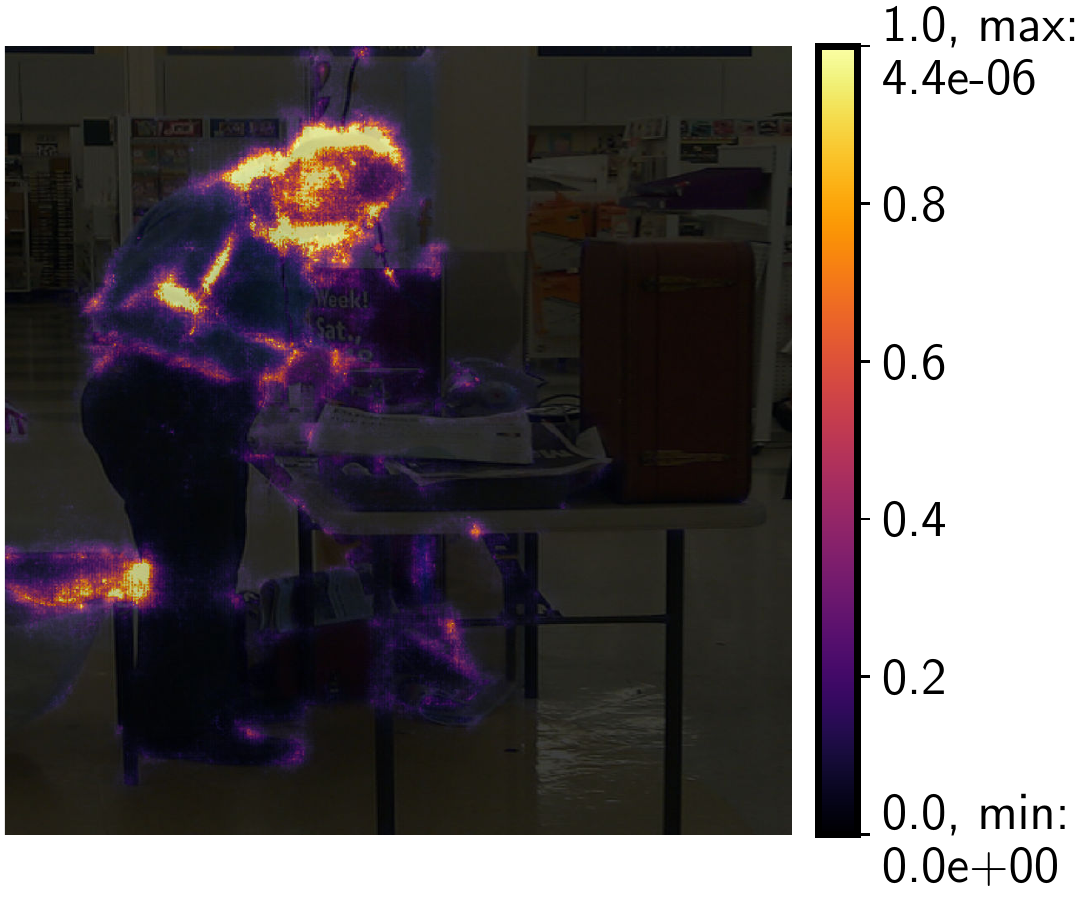}   &
        \includegraphics[scale=0.30,valign=c]{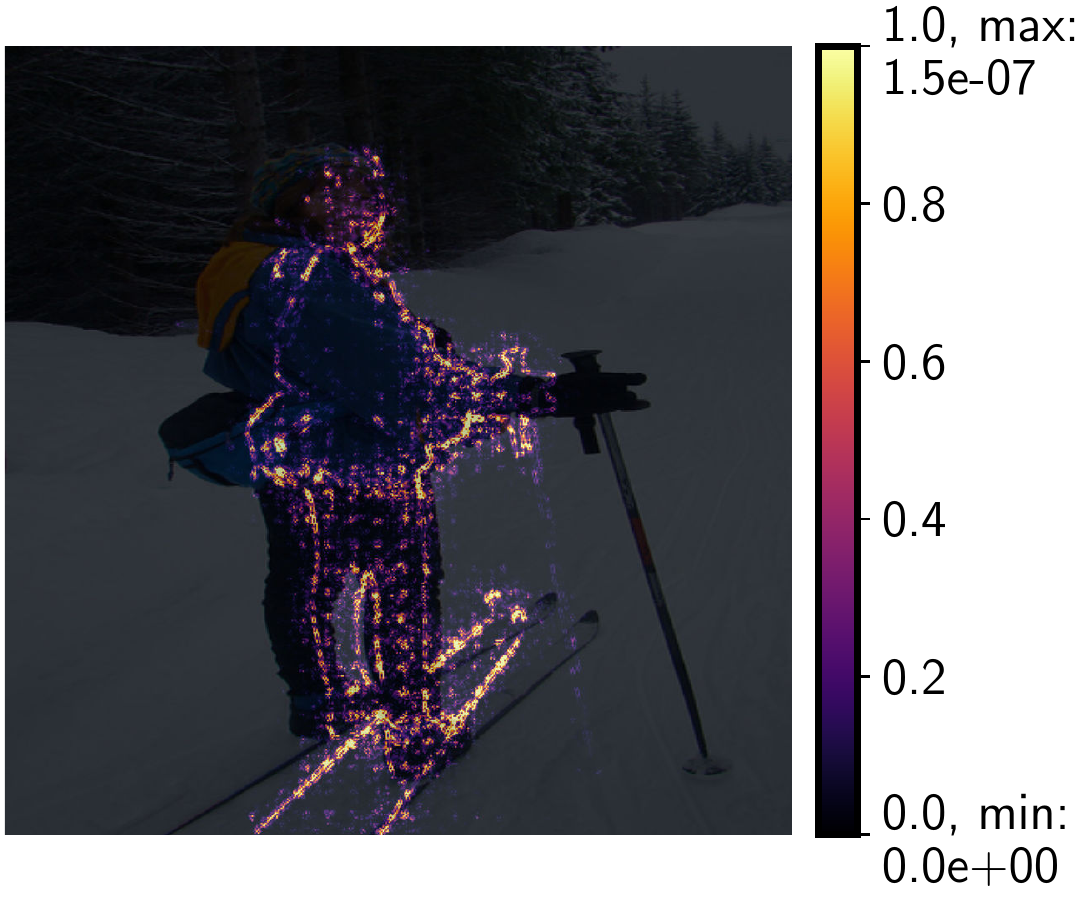}   &
        \includegraphics[scale=0.30,valign=c]{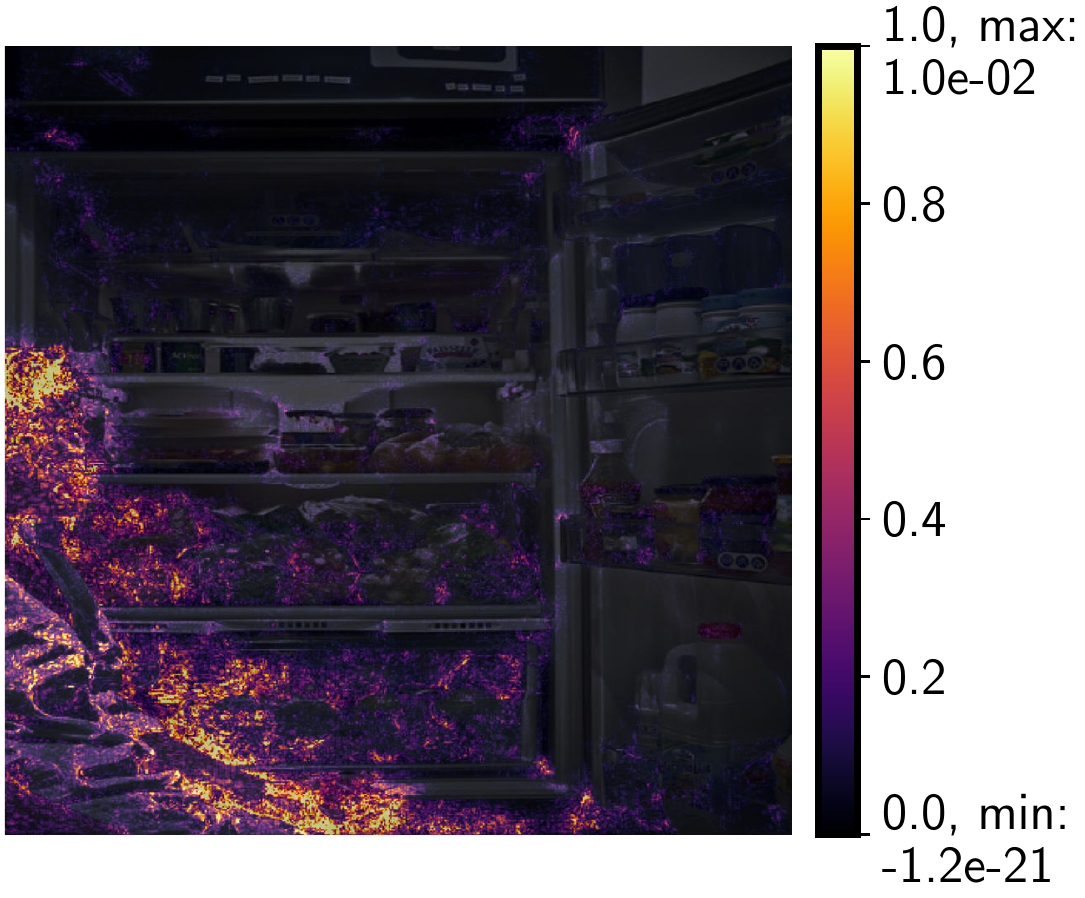}
        \\[1em] 
        
        \rotatebox[origin=c]{90}{Random model} &
        \includegraphics[scale=0.30,valign=c]{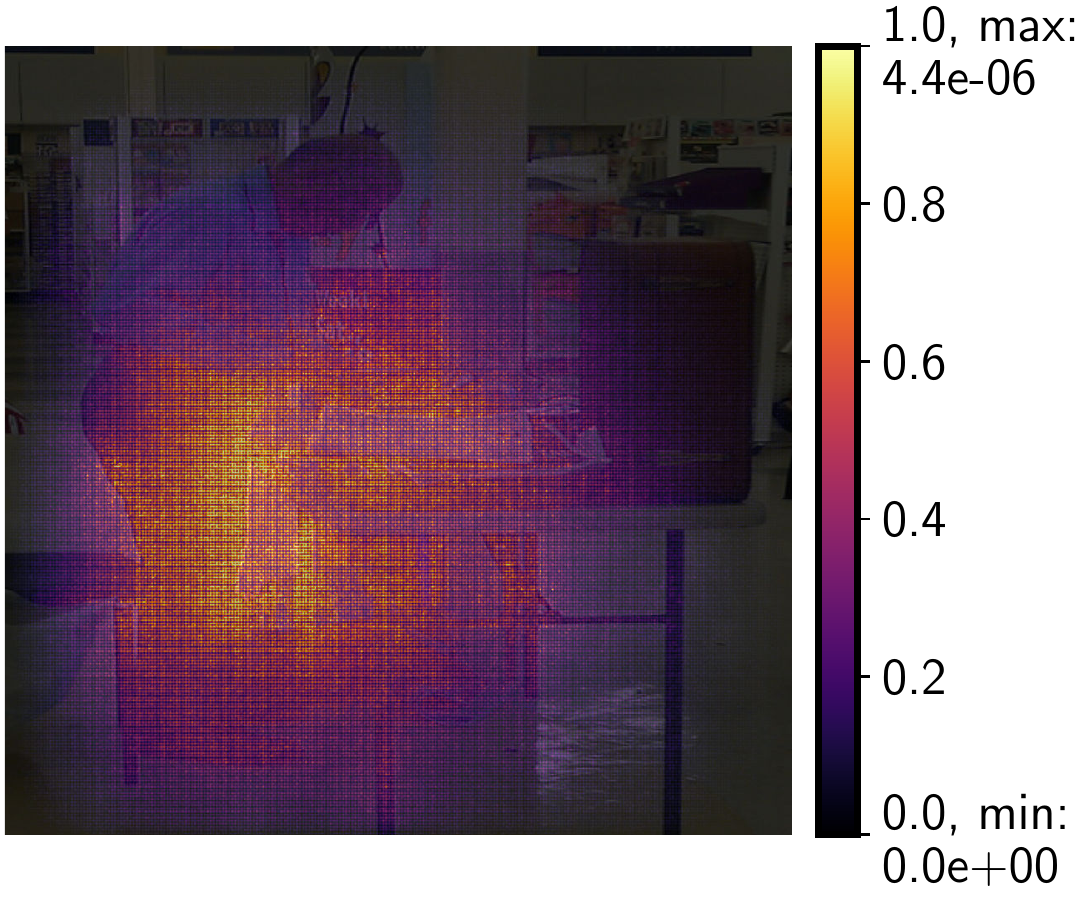}   &
        \includegraphics[scale=0.30,valign=c]{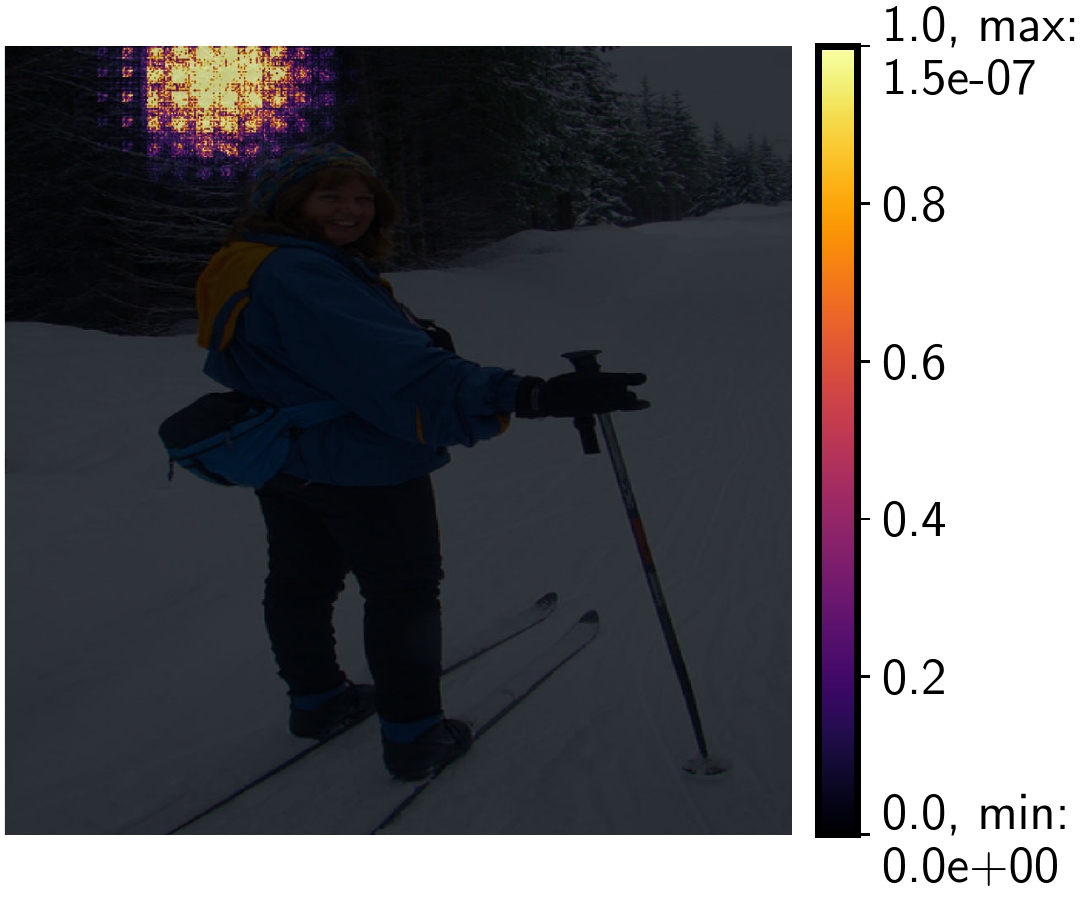}   &
        \includegraphics[scale=0.30,valign=c]{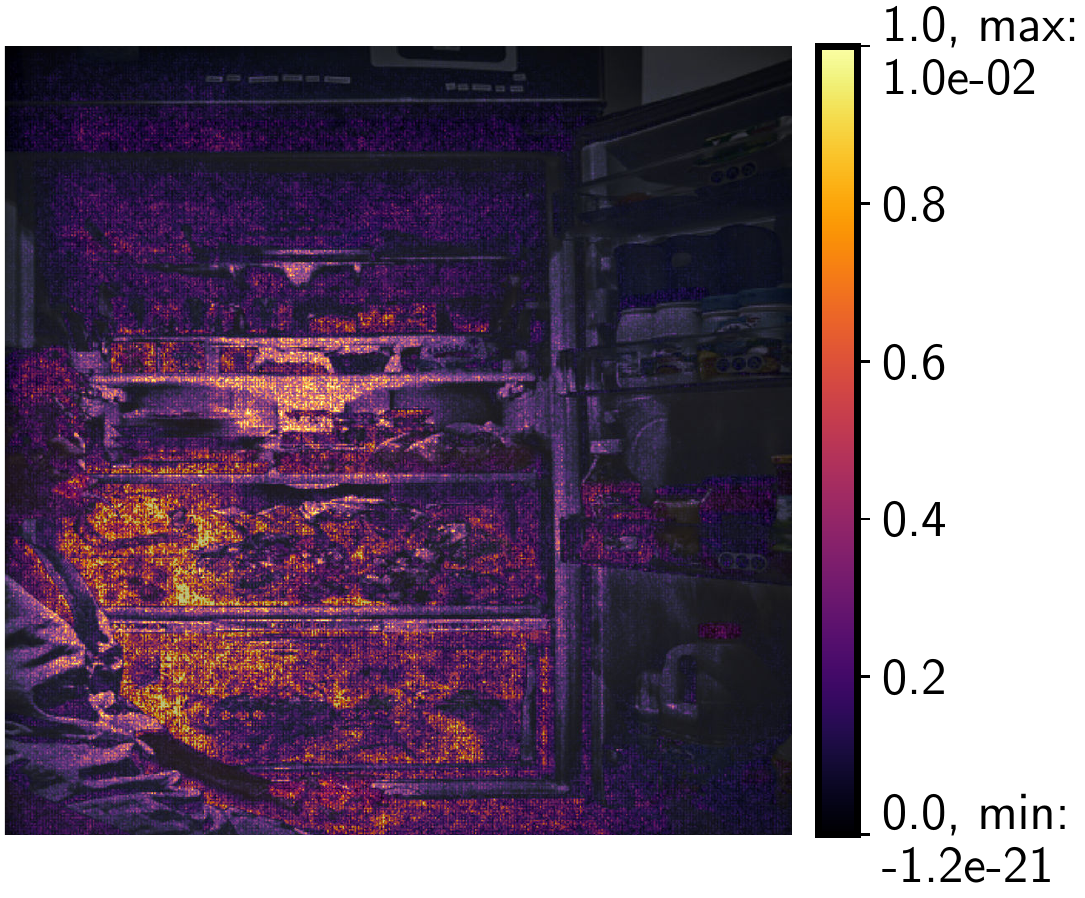}
    \end{tabular}
    \label{fig:sanity_properties}
\end{table*}

\subsection{Quantitative Evaluation Criteria}

For quantitative evaluation, in order to assess the change in saliency maps when randomizing the model parameters, the similarity between the classification decision saliency maps generated from each randomized model instance and the true model is computed using Structural Similarity (SSIM). This allows for visual changes to the saliency map to be compared and tracked.

\subsection{Qualitative Evaluation Criteria}
\label{section:sanity_scores}
This section reports on the subjective analysis carried out to understand the differences in sensitivity of explanation methods across various detectors. 
Table~\ref{tab:subjective_analysis} illustrates clearly that the ability to explain is more model-dependent than the ability of the explanation method to interpret a particular model.\\

A comparison is developed by visually inspecting certain aspects of the saliency map obtained using a completely randomized model and also by comparing it with the saliency map generated using the trained model.
The various aspects considered to indicate the magnitude of sensitivity are provided below with the a scoring guide. 
A visual illustration of these aspects is shown in Table~\ref{fig:sanity_properties}.
In the negative scenarios, the methods are awarded a score -1 $\times$ (score awarded below).
A score 1 is added to the total score if the method scores 1 for any one aspect. This indicates that the method passes the sanity test.

Now we define criteria to evaluate a saliency map made by explaining an object detector output.
\newpage
\begin{enumerate}   
	\item \textbf{Edge detector}.
    Saliency methods sometimes act as an edge detector which does not depend on the input image, which is undesirable \cite{Adebayo_sanitychecks}.
	A method acting as an edge detector is scored -1 because the explanations should be meaningful rather than simply behaving like an edge detector.
	
	\item \textbf{Highlight only interest object}.
    Saliency explanations should be focused on the interest object inside the bounding box, assuming that that model performs adequately and is not fooled by context or background \cite{Rosenfeld_Elephant}. A model with randomized should not have this behavior as information was destroyed and the saliency map should reflect this.
	When the saliency map generated using the randomized model only highlights the interest object explained, the method is awarded a -1 score.
	
	\item \textbf{Focus more than one object}. 
    Opposite from the previous criteria, a randomized model should focus in more than one object as there is no object-specific information in the model.
	Score of 1 is awarded to the method producing a saliency map that highlights more than a single object in the image.
	
	\item \textbf{Texture change}. The texture of a saliency map denotes the spatial arrangement of intensity in a pattern over an image region.
	If the texture of the saliency map obtained using the randomized model varies from that of the saliency map of the true model, the method is awarded a score 1.
	For instance, the randomized model map can be a smoothened version without sharp features or completely hazy.
	
	\item \textbf{Illustrate artifacts}.
    Artifacts in saliency maps are also undesirable as they show bias in the model structure and/or equations which affect the quality of a saliency map.
	If the saliency map from the randomized model displays certain image artifacts such as checkerboard artifacts and sharp parallel lines, the method is awarded -1.
	
	\item \textbf{Intensity range change}.
    The range of pixel values in a saliency map should change as the model is randomized, reflecting the destruction in information when weights are randomized.
	Score of 1 is awarded if the saliency map intensity range changes before normalizing between 0 to 1 across the randomized and true model.
\end{enumerate}

\section{Experimental Setup}

\textbf{Object Detectors}. In this study we evaluate three pre-trained object detectors: Faster R-CNN (FRN) \cite{Ren_FasterRCNN}, SSD512 (SSD) \cite{Liu_SSD}, and EfficientDet-D0 (EDO0) \cite{Tan_EfficientDet}, all trained on the COCO dataset \cite{Lin_MSCOCO}. Detailes are provided in Table~\ref{tab:selected_detectors_detail}.

\begin{table*}[!htb]
	\small
	\centering
	\caption{Summary of object objector implementations used in this work. The detectors are trained to detect common objects using COCO dataset. The mAP reported is at 0.5 IoU threshold. val35k represents 35k COCO validation split images. minival is the remaining images in the validation set after sampling val35k.}
	\begin{tabular}{@{} lllllll @{}}
		\toprule
		& & \multicolumn{2}{c}{\textbf{COCO split}} & \\
		\textbf{Detector} & \textbf{Stage} & \textbf{Train set} & \textbf{Test set} & \textbf{mAP (\%)} & \textbf{Weights} & \textbf{Code} \\
		\midrule
		Faster R-CNN & Two & train+val35k 2014 & minival2014 & 54.4  & \cite{Matterport_MaskRCNN} & \cite{Matterport_MaskRCNN}\\
		SSD512  & Single & train+val35k 2014 & test-dev 2015 & 46.5 & \cite{Liu_SSD} & \cite{Octavio_PAZ}\\
		EfficientDet-D0 & Single & train 2017 & test-dev 2017 & 53.0 & \cite{Tan_EfficientDet} & \cite{Octavio_PAZ} \\		
		\bottomrule
	\end{tabular}
	\label{tab:selected_detectors_detail}
\end{table*}

\begin{table}[!htb]
	\centering
	\caption{Details about the marine debris objector used in this work. The mAP reported is at 0.5 IoU threshold.}
	\begin{tabular}{lll} 
		\toprule
		\textbf{SSD Backbones} & \textbf{mAP (\%)} & \textbf{Input Image Size} \\ 
		\midrule
		VGG16 & 91.69 & 300 x 300 \\
		ResNet20 & 89.85 & 96 x 96 \\
		MobileNet & 70.30 & 96 x 96 \\
		DenseNet121 & 73.80 & 96 x 96 \\
		SqueezeNet & 68.37 & 96 x 96 \\
		MiniXception & 71.62 & 96 x 96 \\
		\bottomrule
	\end{tabular}
	\label{tab:selected_marine_debris_detectors_detail}
\end{table}

\textbf{Explanation Methods}. We evaluate several gradient-based saliency methods, namely Guided Backpropagation (GBP) and Integrated Gradients (IG), as well as their variations using SmoothGrad (SGBP and SIG). Mathematical details for these methods are provided in the appendix. 

\textbf{Datasets}. The detectors trained on common objects are used to perform the model randomization test. 
The detector details are available in Table~\ref{tab:selected_detectors_detail}.
Therefore, the model randomization test is carried out for all the 12 combinations of detectors and explanation methods. 
The dataset used for the model randomization study is the COCO test 2017 split \cite{Lin_MSCOCO}. 
15 randomly sampled images from the COCO test 2017 split is analyzed for model randomization test.
The test split is chosen because the train and validation splits are used in training the detectors.

In order to perform the data randomization test, the Marine Debris dataset \cite{Matias_MarineDebris} \cite{Matias_MarineDebris_Dataset} is used. 
This study uses two versions of SSD trained on the Marine Debris dataset.  Details and performance of detectors trained on Marine Debris Dataset are shown in Table~\ref{tab:selected_marine_debris_detectors_detail}.
The two versions are true and random SSD models with VGG16 backbone trained using the true and random labels respectively.
The additional details about the true SSD-VGG16 model is provided in Table~\ref{tab:selected_marine_debris_detectors_detail}.
The random detector is trained using random class labels and adding random noise to the ground truth box coordinates.
The random detector is trained until the mAP@[IoU=0.5] on the train set is 80\%. 
The explanations are generated for the test set images.
The Marine Debris dataset is used for this experiment to overcome the time taken to train a detector on a complex COCO dataset.
In addition, the Marine debris dataset aids in studying the applicability of explaining detectors in a real-world application.

\FloatBarrier

\begin{figure*}[!htb]
	\centering
	\mbox{} \hfill
	\includegraphics[width=\linewidth]{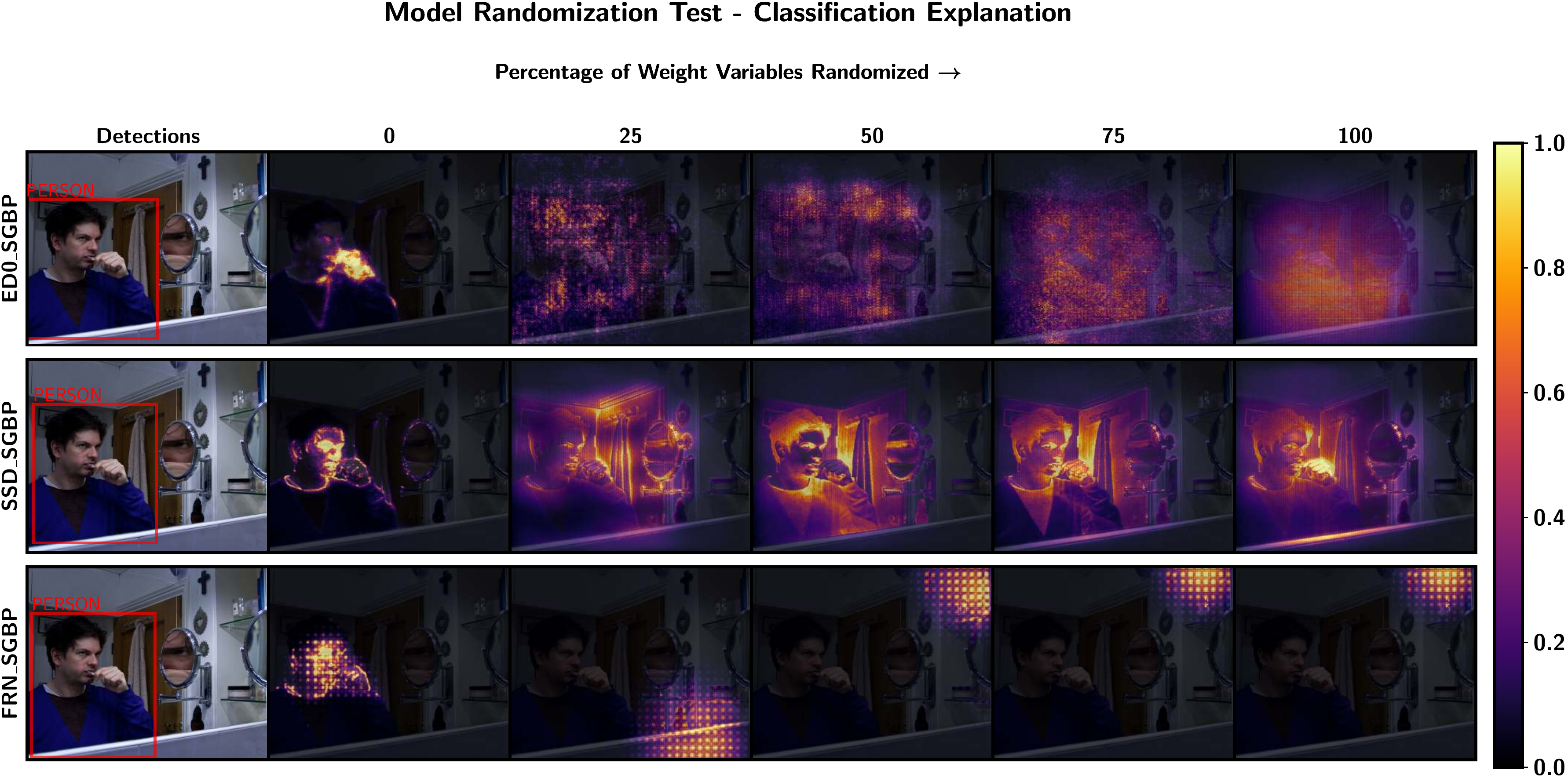}
	\caption{\label{fig:sanity_class_sgbp} Model randomization test for classification explanations (red-colored box) across different models using SGBP. The first column is the detection of interest that is explained in the consecutive columns. 
		The second column is the saliency map generated using the trained model without randomizing any parameters, which highlights the important parts such as hands, eyes, and face. The last column is the saliency map generated using a model with all parameters randomized. Note how FRN fails the randomization test.
	}
\end{figure*}

\begin{figure*}[!htb]
	\centering
	\mbox{} \hfill
	\includegraphics[width=\linewidth]{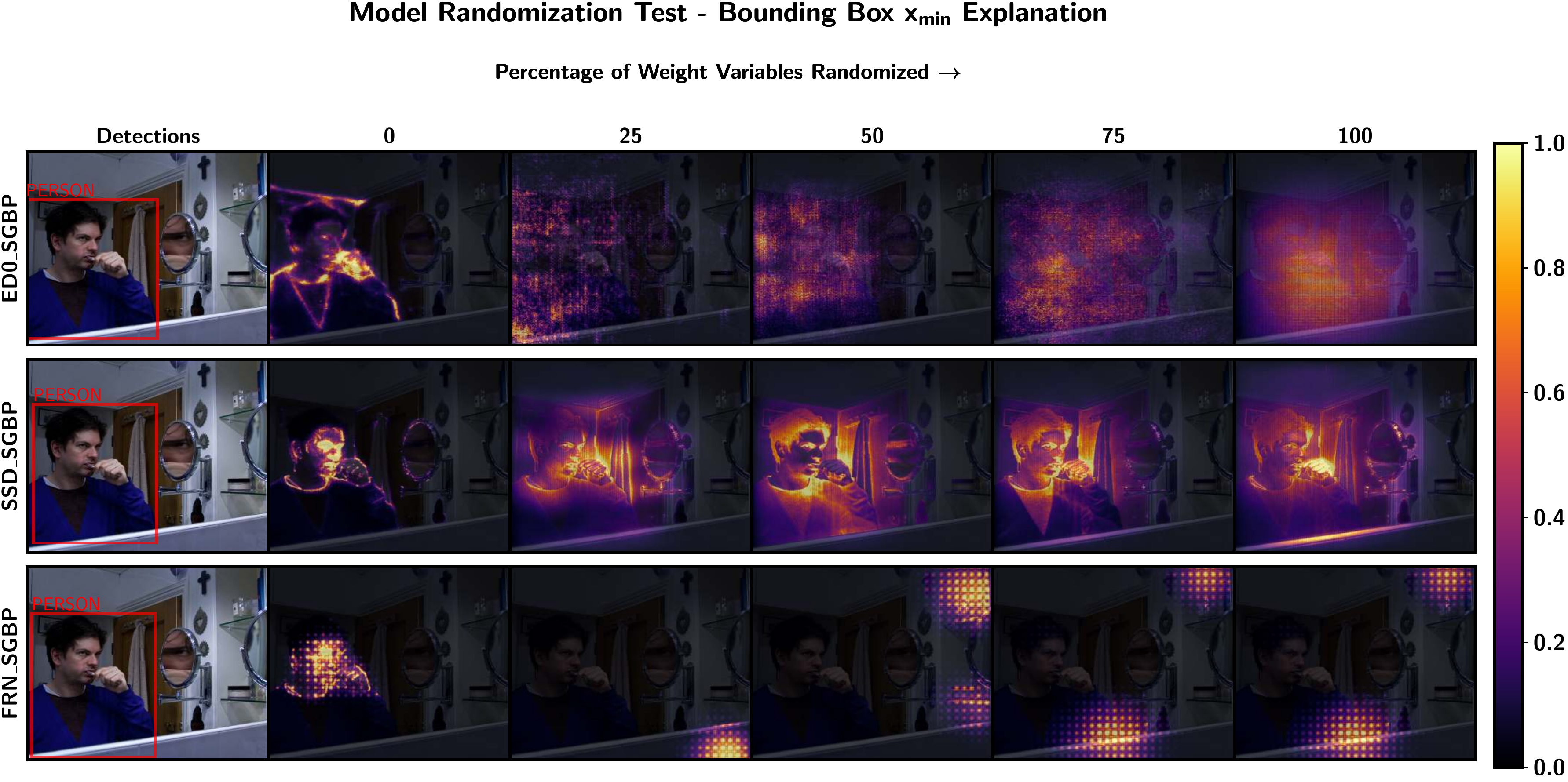}
	\caption{\label{fig:sanity_xmin_sgbp} Model randomization test for $x_{min}$ explanations (red-colored box) across different models using SGBP. The first column is the detection of interest that is explained in the consecutive columns. 
		The second column is the saliency map generated using the trained model without randomizing any parameters. 
		The second column highlights the important parts such as hands, eyes, and face. 
		The last column is the saliency map generated using a model with all parameters randomized. Note how FRN fails the randomization test.
	}
\end{figure*}

\section{Results and Discussion}

\begin{figure*}[!htb]
	\centering
	\mbox{} \hfill
	\includegraphics[width=\linewidth]{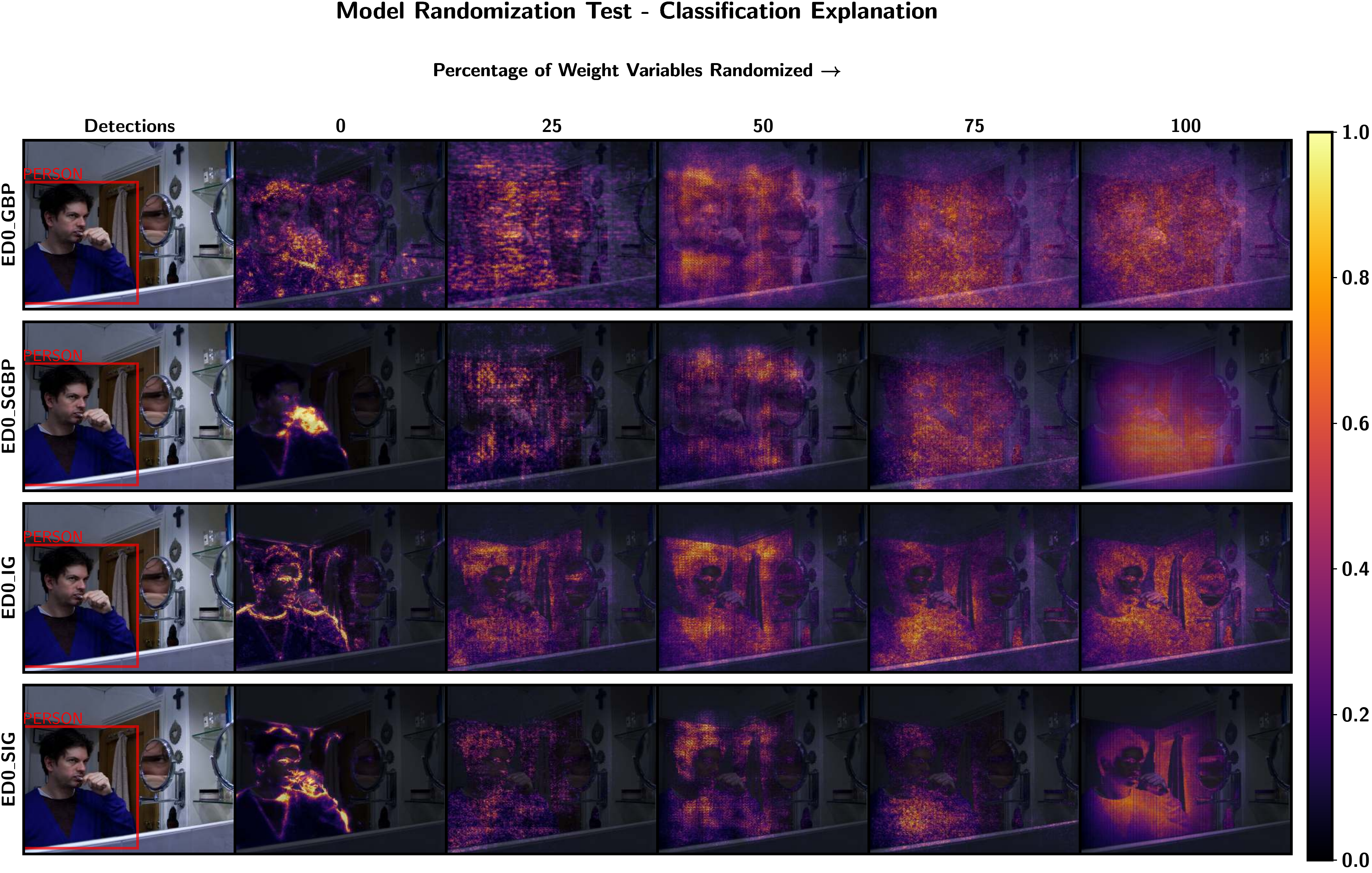}
	\caption[EfficientDet-D0 randomization test across explanation methods for classification decision]{\label{fig:sanity_efficientdet_allmethods_class}Comparison of explanations using different explanation methods for the classification decision corresponding to EfficientDet-D0 detections. 
		The first column is the detection (red-colored box) explained by the methods. 
		The second column is the saliency map generated using the trained model without randomizing any parameters. 
		The last column is the saliency map generated using the model with all parameters randomized. 
		SIG after randomizing 75 percentage of the weight variables visually highlight certain regions of the person detection. 
		However, the magnitude is relatively very less and texture of the map is considerably different to the true model explanation.
	}
\end{figure*}

\textbf{Model randomization test:} The saliency maps are investigated for both the bounding box and classification decisions corresponding to a detection.
The model parameter randomization randomizes the weight variables starting from the last layers.
The left-most column after the interest detection with 0\% represents the saliency map generated using the trained model with none of the weight variables randomized. 
100\% in the last column is the saliency maps generated using a randomly initialized model with all weight variables completely randomized. 
Figure~\ref{fig:sanity_class_sgbp} illustrates the effect of classification explanations to the model parameters. 
In the case of the EfficientDet-D0 classification explanation with SGBP, the saliency map is completely noisy without highlighting any specific feature. 
SSD with SGBP acts like an edge detector by sharply highlighting certain features as the number of weight variables randomized changes.
However, the saliency map highlights feature other than the person object. 
Figure~\ref{fig:sanity_xmin_sgbp} illustrate the sensitivity of box coordinate $x_{min}$ explanations using SGBP to the model characteristics. 
The saliency maps highlight regions of the person at a certain randomization level for SIG as shown in Figure~\ref{fig:sanity_efficientdet_allmethods_class}.

The magnitude of change between the saliency maps of true and randomized model is different for each model as the weight variables are randomized.
It clearly illustrates model randomization tests should be performed for each model and method combinations as stated in the related work.
Section~\ref{section:sanity_scores} discusses subjectively the magnitude of the change in sensitivity across detectors and explanation method combinations.
Therefore, the ability to explain models are more dependent on the model than the ability of the explanation method to explain the model.

The explanation using GBP for EfficientDet-D0 is noisy because the GBP method acts similar to the Gradients method in the case of EfficientDet-D0. Gradients estimate the gradient of the output target neuron with the input. Since there are no ReLU activations for EfficientDet-D0 the negative contributions are not retarded and the prime usage of GBP is relaxed to work as Gradients method.

\begin{figure*}[!t]
	\begin{minipage}[c][0.8\width]{
			0.3\textwidth}
		\centering
		\includegraphics[width=1.0\textwidth]{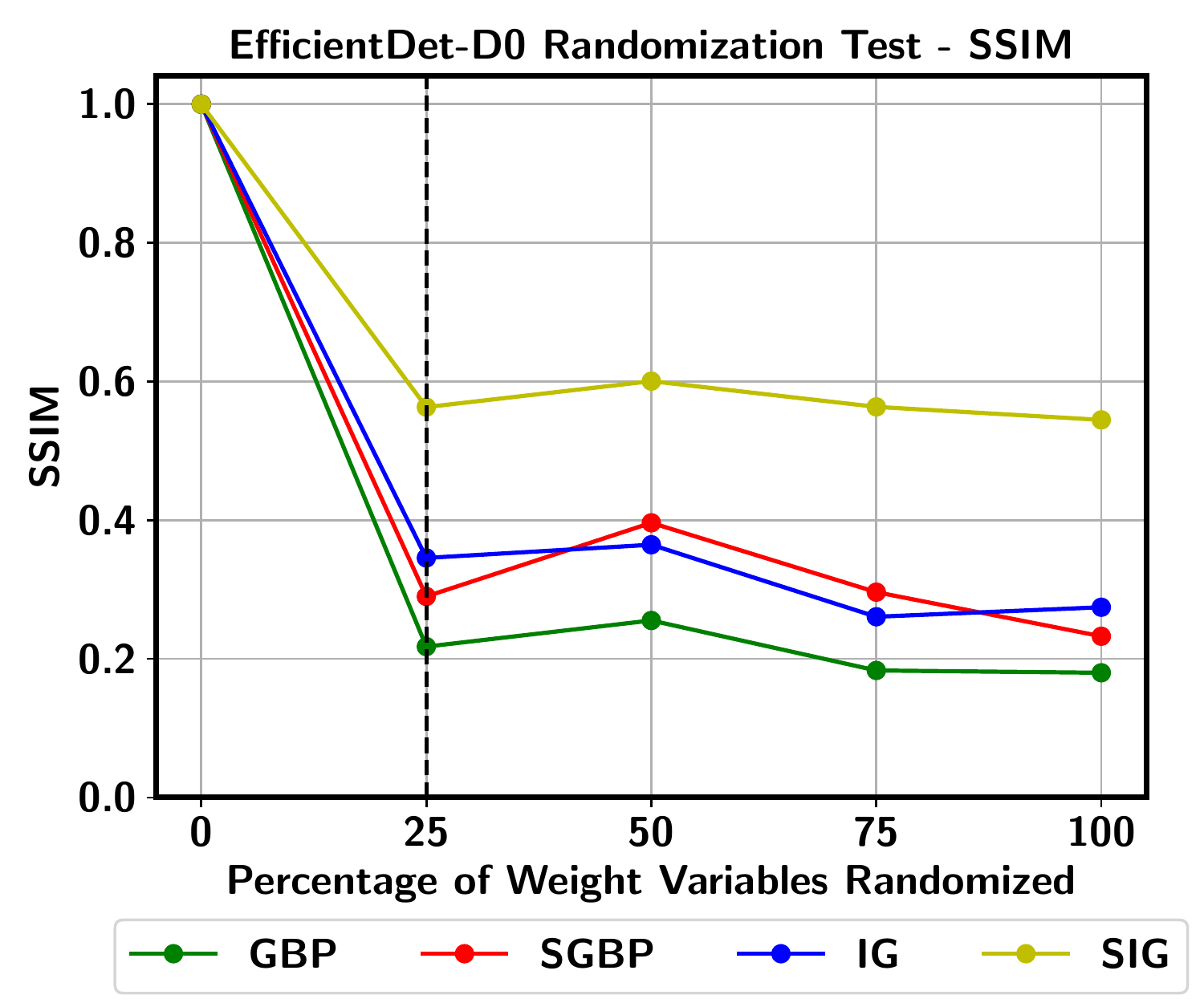}
	\end{minipage}
	\hfill 	
	\begin{minipage}[c][0.8\width]{
			0.3\textwidth}
		\centering
		\includegraphics[width=1.0\textwidth]{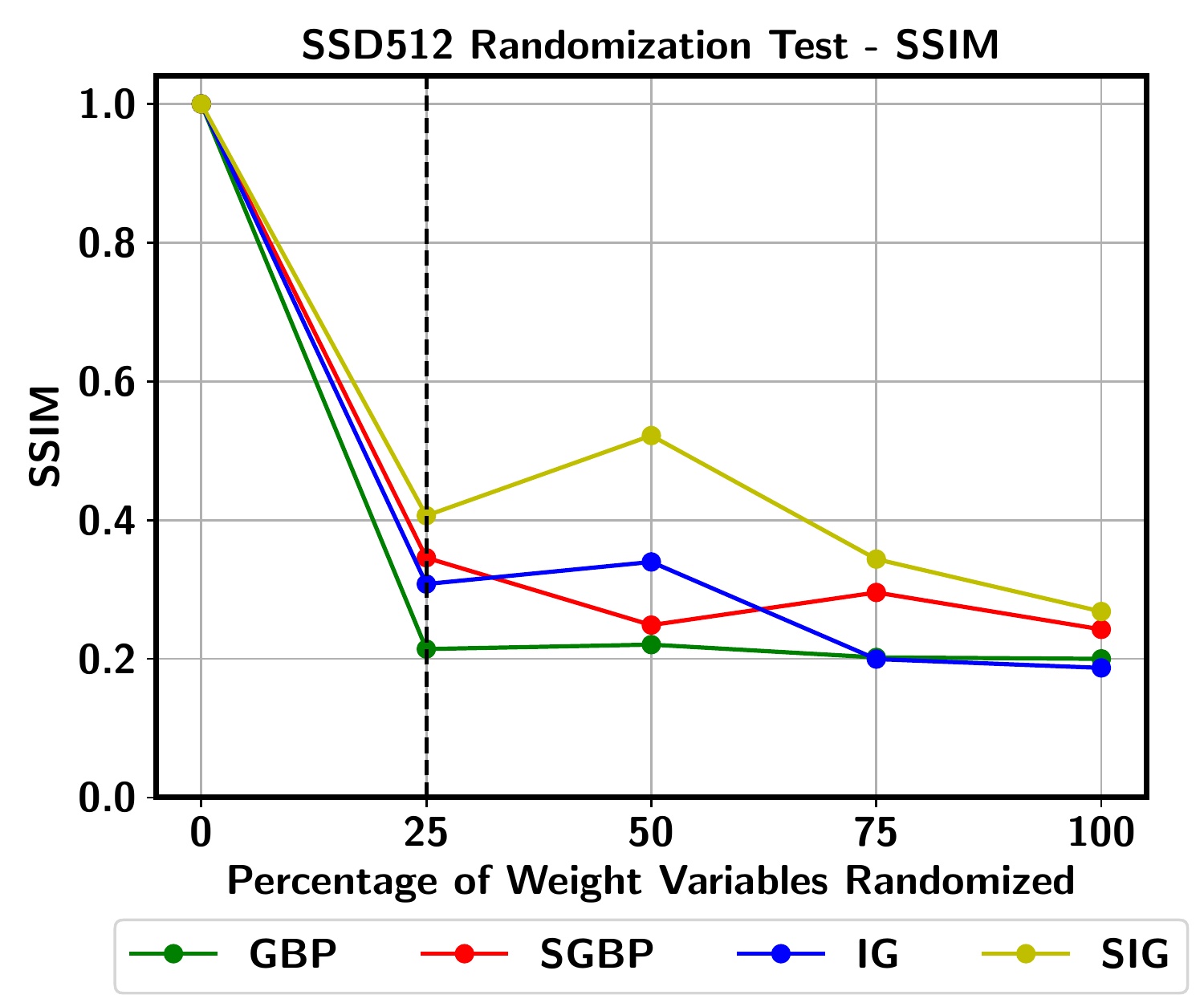}
	\end{minipage}
	\hfill
	\begin{minipage}[c][0.8\width]{
			0.3\textwidth}
		\centering
		\includegraphics[width=1.0\textwidth]{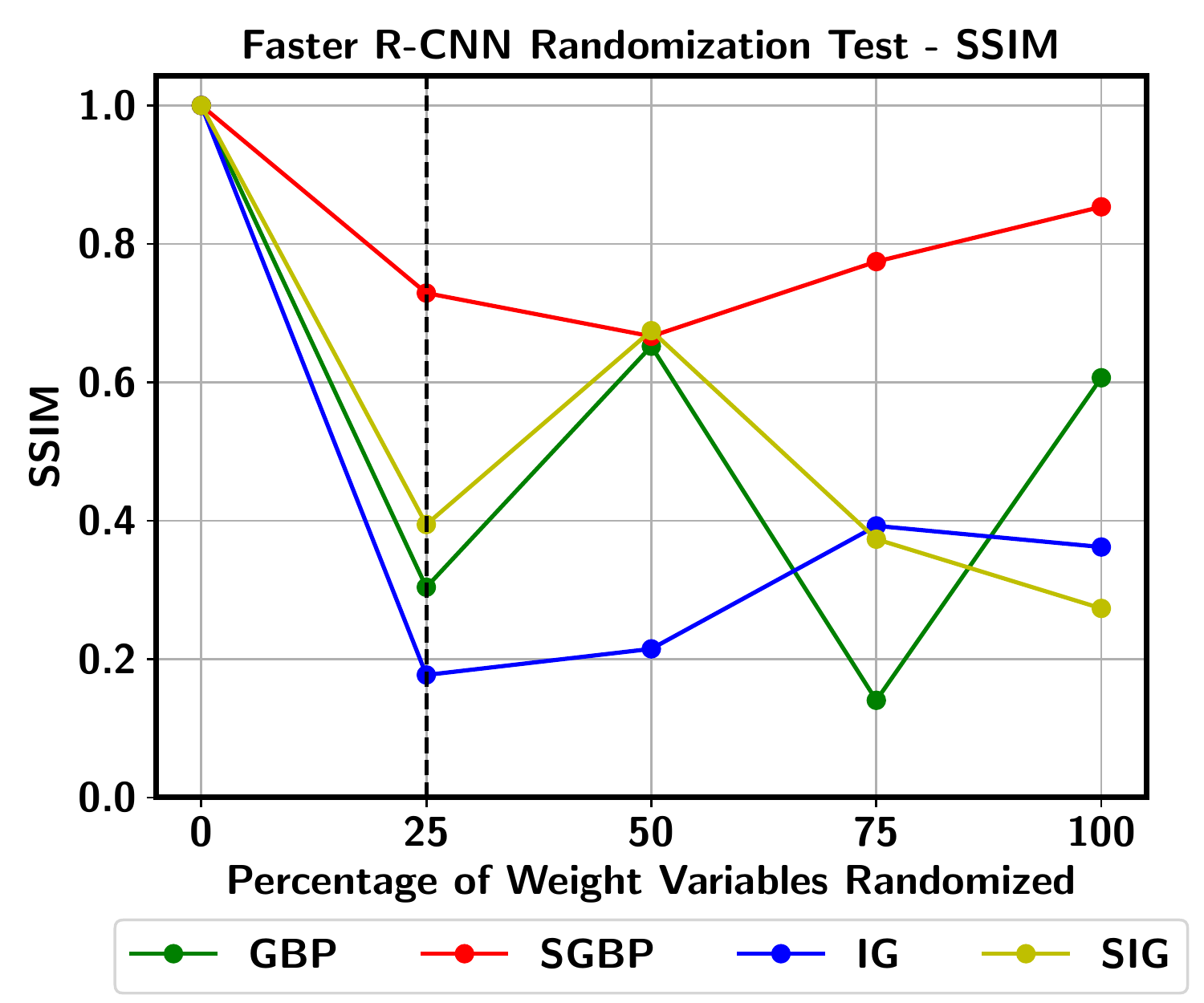}
	\end{minipage}
	\caption[A quantitative assessment of model randomization test using SSIM]{A quantitative assessment using SSIM of the change in classification saliency map features during model randomization test across explanation methods and detectors is provided. SSIM is the average SSIM computed across a subset of test images sampled from the COCO test 2017.}
	\label{fig:ssim_graphs}
\end{figure*}

The SSIM in Figure~\ref{fig:ssim_graphs} is the average SSIM across different percentage of weight variables randomized for a set of 15 images randomly sampled from the COCO test set.
Since the explanations have changed in terms of the important pixels highlighted, saliency map texture, and SSIM metric with regards to the explanations using the true model, all the explanation methods pass the model randomization test for detectors.

The gradient attribution maps for the two-stage detector, Faster R-CNN, illustrate checkerboard artifact on randomizing weights as shown in Figure~\ref{fig:sanity_class_sgbp}. 
There are various reasons for the gradient artifacts as discussed in \cite{Odena_artifacts} \cite{Sugawara_artifacts}. 

In the case of using GBP and SGBP with Faster R-CNN, the higher SSIM between the classification decision saliency maps of completely randomized model and true model is because of the checkerboard artifact shown in Figure~\ref{fig:sanity_class_sgbp}.
Even though the center of mass of the grid pattern shifts over the image, the SSIM provides a higher score due to similarity in the pattern. 
This observation is in agreement with the subjective analysis in Section~\ref{section:sanity_scores} with low sensitivity scores for Faster R-CNN - GBP as well as Faster R-CNN - SGBP compared to other detector and explanation method combinations.

\begin{figure}[t!]
	\centering
	\includegraphics[width=0.9\linewidth]{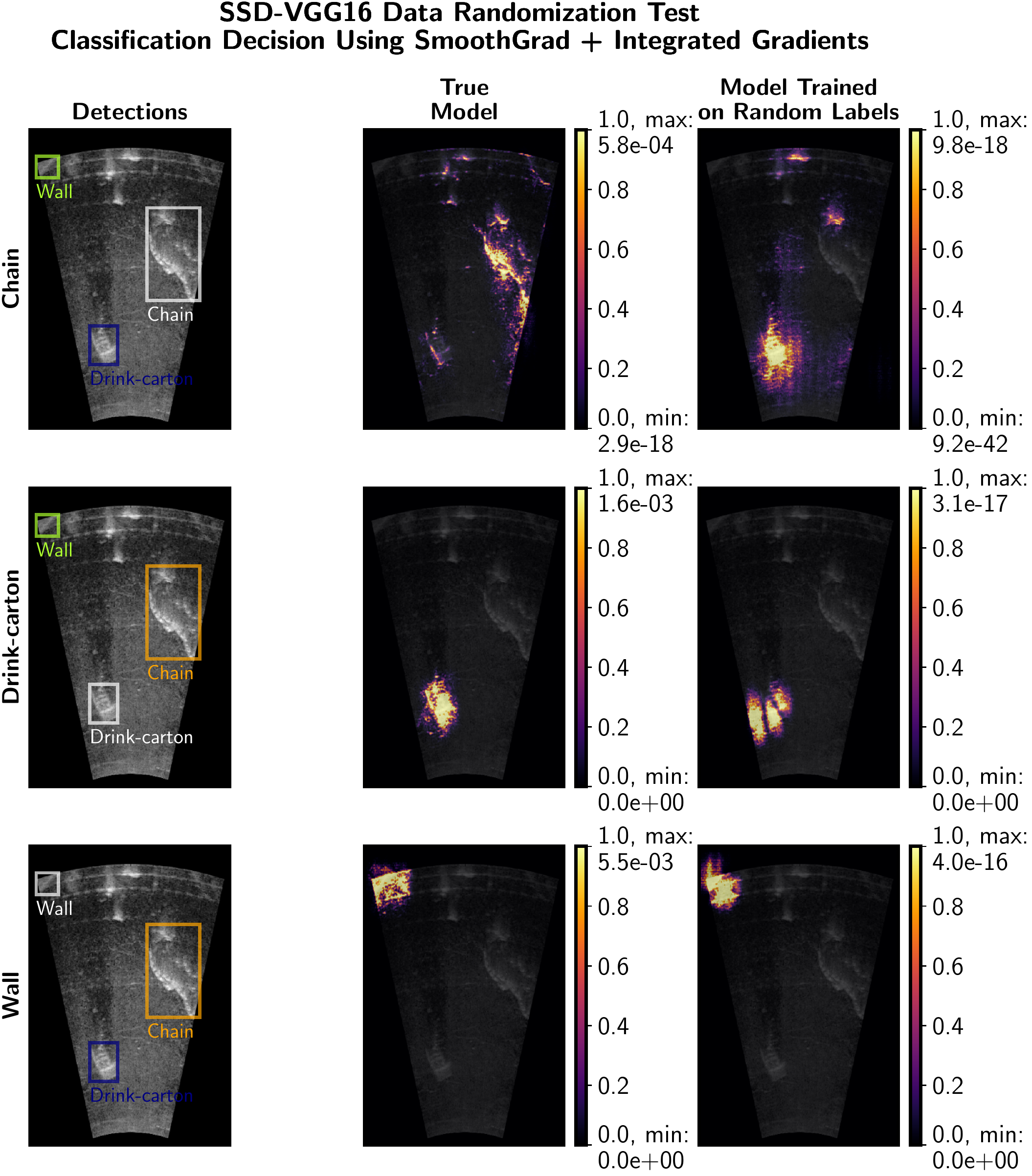}
	\caption{\label{fig:data_randomization_test_ex1} Data randomization test using SSD-VGG16 and SIG. The saliency maps explains the classification decision. The first column depicts the detections, the detection of interest is highlighted in white. The true and random model classification explanations differ in terms of the features highlighted, attribution intensity, and the explanation texture.
	}
\end{figure}

\begin{figure}[t!]
	\centering
	\includegraphics[width=0.9\linewidth]{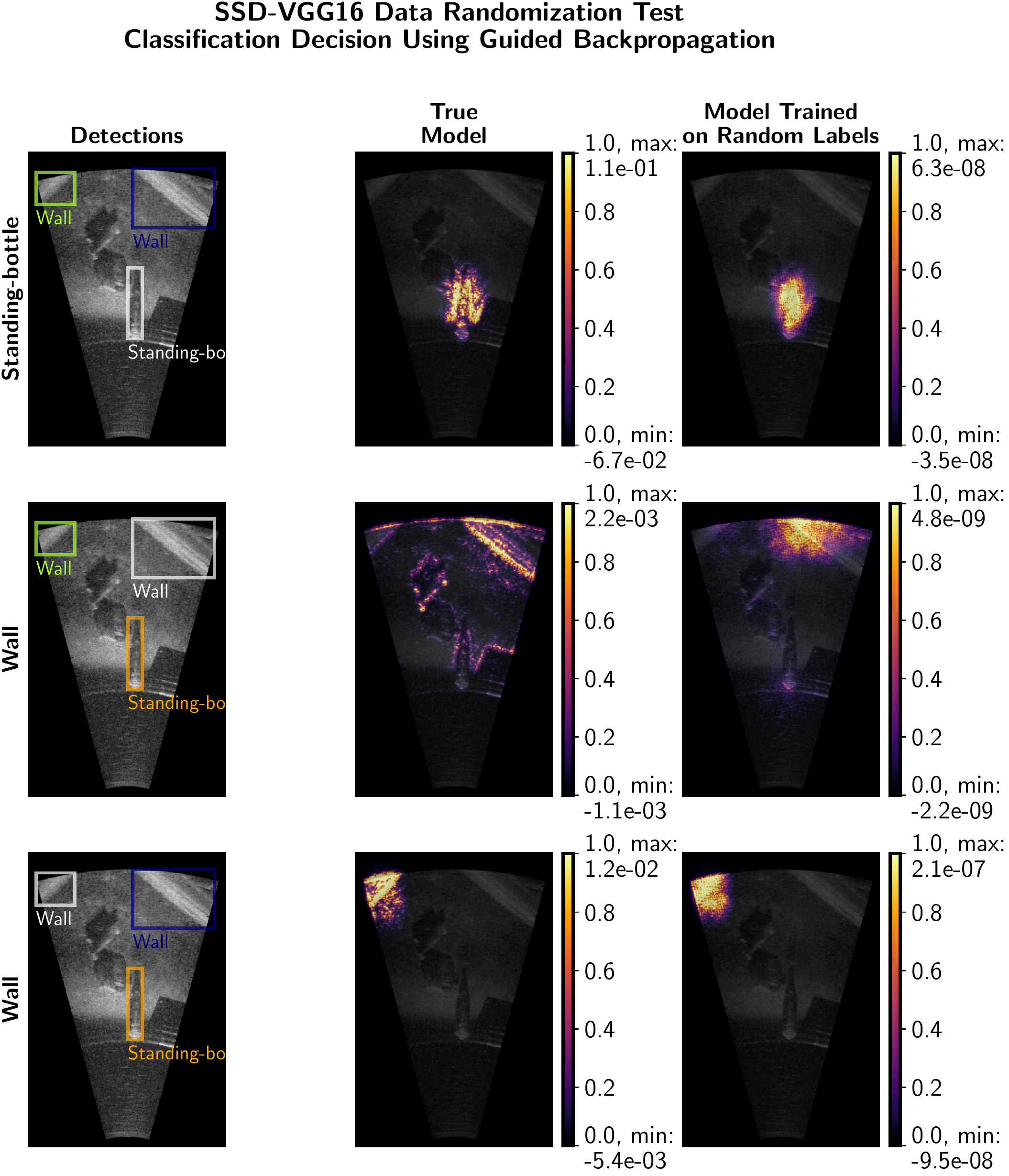}
	\caption{\label{fig:data_randomization_test_ex2} Data randomization test using SSD-VGG16 and GBP. The saliency maps explains the classification decision. The first column depicts the detections, the detection of interest is highlighted in white. The true and random model classification explanations differ in terms of the features highlighted, attribution intensity, and the explanation texture.
	}
    \vspace*{8em}
\end{figure}

\textbf{Data randomization test:} 
Figure~\ref{fig:data_randomization_test_ex1} illustrate the differences in the saliency maps explaining classification decision of SSD-VGG16. 
The attribution map intensity levels are largely different. 
The texture of the explanations from the random model looks smoothed. 
However, the explanations generated using the true model illustrate sharp features.
There are substantial differences in the saliency map generated using SIG for the chain detection in Figure~\ref{fig:data_randomization_test_ex1}.
In addition, the drink-carton classification explanations for the random model illustrates patches, where as, the drink-carton is relatively sharper in the explanation from the true model.
However, the difference for other detection is only at the level of attribution intensity and texture. 
Therefore, this opens up the possibility to perform sanity checks at the class level. 
The methods should remain sensitive for each class predicted by the model.
The findings is consistent in Figure~\ref{fig:data_randomization_test_ex2} for explanations generated using GBP.
In addition, all the explanation methods provide different saliency maps for both classification and bounding box explanations in terms of features highlighted, saliency map texture, and the attribution intensity.
Therefore, none of the selected explanation methods fail the data randomization test for the SSD-VGG16 detector. 

\section{Conclusions and Future Work}

In this work we have evaluated standard sanity checks for saliency explanations in object detectors, considering both object classification and bounding box regression explanations, through data and weight randomization. We defined new qualitative criteria to systematically evaluate saliency maps visually, and we find that overall, more modern object detectors like EfficientDet-D0 pass more sanity checks and provide higher quality saliency explanations than older detectors like SSD and Faster R-CNN.

Our conclusions hold under multiple gradient-based saliency methods, we tested Guided Backpropagation and Integrated gradients, as well as their Smooth Gradient combinations.

When Faster R-CNN fails to be explained using gradient-based saliency maps, there are large checkerboard artifacts in the explanation, which stay even as weights are randomized. SSD does not produce checkboard patterns but the explanation is insensitive to weights being randomized. Only EfficientDet-D0 produces explanations that pass both data and weight randomization checks.

We expect that our work can increase interest in object detector explanations, and provide additional ways to empirically validate these explanations. We believe that our work provides additional insights not covered by \cite{Adebayo_sanitychecks}, specially using multiple and more complex models like object detectors.

\textbf{Limitations}. On a broader note, our work can focus on a larger evaluation set with more defined evaluation metrics to assess the saliency maps. The evaluation set is limited due to the high computation time to generate saliency maps for each detection in an image for all coordinates, category decision, and randomization levels. 

In addition, we consider that certain models are not explainable based solely on the fact that a few explanation methods fail in effectively explaining certain detector decision. To make informed decisions, more explanation methods should be evaluated together with sanity checks. Our work only provides a limited view on this problem, but we do show that explainability depends both on model and saliency explanation method.

\clearpage
\newpage
\bibliographystyle{splncs04}
\bibliography{report_bibliography.bib}

\begin{thebibliography}{10}
\providecommand{\url}[1]{\texttt{#1}}
\providecommand{\urlprefix}{URL }
\providecommand{\doi}[1]{https://doi.org/#1}

\bibitem{Matterport_MaskRCNN}
Abdulla, W.: {Mask R-CNN for object detection and instance segmentation on
  Keras and TensorFlow}. GitHub (2017), (Online accessed on 20 September 2021)

\bibitem{Adebayo_sanitychecks}
Adebayo, J., Gilmer, J., Muelly, M., Goodfellow, I.J., Hardt, M., Kim, B.:
  {Sanity Checks for Saliency Maps}. In: Bengio, S., Wallach, H.M., Larochelle,
  H., Grauman, K., Cesa{-}Bianchi, N., Garnett, R. (eds.) Advances in Neural
  Information Processing Systems ({NeurIPS}). vol.~31, pp. 9525--9536. Curran
  Associates, Inc. (2018)

\bibitem{Arani_ODSurvey}
Arani, E., Gowda, S., Mukherjee, R., Magdy, O., Kathiresan, S.S., Zonooz, B.: A
  comprehensive study of real-time object detection networks across multiple
  domains: A survey. Transactions on Machine Learning Research  (2022), survey
  Certification

\bibitem{Araujo_UOLO}
Ara{\'{u}}jo, T., Aresta, G., Galdran, A., Costa, P., Mendon{\c{c}}a, A.M.,
  Campilho, A.: {{UOLO} - Automatic Object Detection and Segmentation in
  Biomedical Images}. In: Stoyanov, D., Taylor, Z., Carneiro, G.,
  Syeda{-}Mahmood, T.F., Martel, A.L., Maier{-}Hein, L., Tavares, J.M.R.S.,
  Bradley, A.P., Papa, J.P., Belagiannis, V., Nascimento, J.C., Lu, Z.,
  Conjeti, S., Moradi, M., Greenspan, H., Madabhushi, A. (eds.) Deep Learning
  in Medical Image Analysis and Multimodal Learning for Clinical Decision
  Support. {DLMIA} 2018 and {ML-CDS} 2018. Lecture Notes in Computer Science
  ({LNCS}), vol. 11045, pp. 165--173. Springer, Cham (2018)

\bibitem{Octavio_PAZ}
Arriaga, O., Valdenegro{-}Toro, M., Muthuraja, M., Devaramani, S., Kirchner,
  F.: {Perception for Autonomous Systems {(PAZ)}}. Computing Research
  Repository ({CoRR})  \textbf{abs/2010.14541} (2020)

\bibitem{Bach_LRP}
Bach, S., Binder, A., Montavon, G., Klauschen, F., M{\"{u}}ller, K., Samek, W.:
  {On Pixel-Wise Explanations for Non-Linear Classifier Decisions by Layer-Wise
  Relevance Propagation}. PLOS ONE  \textbf{10}(7),  1--46 (07 2015)

\bibitem{Feng_ADsurvey}
Feng, D., Haase{-}Sch{\"{u}}tz, C., Rosenbaum, L., Hertlein, H., Gl{\"{a}}ser,
  C., Timm, F., Wiesbeck, W., Dietmayer, K.: {Deep Multi-Modal Object Detection
  and Semantic Segmentation for Autonomous Driving: Datasets, Methods, and
  Challenges}. {IEEE} Transactions on Intelligent Transportation Systems
  ({TITS})  \textbf{22}(3),  1341--1360 (2021)

\bibitem{Gudovskiy_E2X}
Gudovskiy, D.A., Hodgkinson, A., Yamaguchi, T., Ishii, Y., Tsukizawa, S.:
  {Explain to Fix: {A} Framework to Interpret and Correct {DNN} Object Detector
  Predictions}. Computing Research Repository ({CoRR})  \textbf{abs/1811.08011}
  (2018)

\bibitem{He_TextDetector}
He, P., Huang, W., He, T., Zhu, Q., Qiao, Y., Li, X.: {Single Shot Text
  Detector with Regional Attention}. In: 2017 {IEEE} International Conference
  on Computer Vision ({ICCV}). pp. 3066--3074. Institute of Electrical and
  Electronics Engineers ({IEEE}) (2017)

\bibitem{Quantus}
Hedström, A., Weber, L., Krakowczyk, D., Bareeva, D., Motzkus, F., Samek, W.,
  Lapuschkin, S., Höhne, M.M.C.: Quantus: An explainable ai toolkit for
  responsible evaluation of neural network explanations and beyond. Journal of
  Machine Learning Research  \textbf{24}(34),  1--11 (2023)

\bibitem{Huang_aisafety}
Huang, X., Kroening, D., Ruan, W., Sharp, J., Sun, Y., Thamo, E., Wu, M., Yi,
  X.: A survey of safety and trustworthiness of deep neural networks:
  Verification, testing, adversarial attack and defence, and interpretability.
  Comput. Sci. Rev.  \textbf{37},  100270 (2020)

\bibitem{Kim_sanity_simulations}
Kim, J.S., Plumb, G., Talwalkar, A.: Sanity simulations for saliency methods.
  In: Chaudhuri, K., Jegelka, S., Song, L., Szepesv{\'{a}}ri, C., Niu, G.,
  Sabato, S. (eds.) International Conference on Machine Learning, {ICML} 2022,
  17-23 July 2022, Baltimore, Maryland, {USA}. Proceedings of Machine Learning
  Research, vol.~162, pp. 11173--11200. {PMLR} (2022)

\bibitem{Kindermans_unreliability}
Kindermans, P., Hooker, S., Adebayo, J., Alber, M., Sch{\"{u}}tt, K.T.,
  D{\"{a}}hne, S., Erhan, D., Kim, B.: {The (Un)reliability of Saliency
  Methods}. In: Samek, W., Montavon, G., Vedaldi, A., Hansen, L.K.,
  M{\"{u}}ller, K. (eds.) Explainable {AI:} Interpreting, Explaining and
  Visualizing Deep Learning, Lecture Notes in Computer Science ({LNCS}), vol.
  11700, pp. 267--280. Springer, Cham (2019)

\bibitem{Lin_MSCOCO}
Lin, T., Maire, M., Belongie, S.J., Hays, J., Perona, P., Ramanan, D.,
  Doll{\'{a}}r, P., Zitnick, C.L.: {Microsoft {COCO:} Common Objects in
  Context}. In: Fleet, D.J., Pajdla, T., Schiele, B., Tuytelaars, T. (eds.)
  Computer Vision -- {ECCV} 2014. Lecture Notes in Computer Science ({LNCS}),
  vol.~8693, pp. 740--755. Springer, Cham (2014)

\bibitem{Liu_SSD}
Liu, W., Anguelov, D., Erhan, D., Szegedy, C., Reed, S.E., Fu, C., Berg, A.C.:
  {{SSD:} Single Shot MultiBox Detector}. In: Leibe, B., Matas, J., Sebe, N.,
  Welling, M. (eds.) Computer Vision -- {ECCV} 2016. Lecture Notes in Computer
  Science ({LNCS}), vol.~9905, pp. 21--37. Springer, Cham (2016)

\bibitem{Marcinkevics_IMLZoo}
Marcinkevics, R., Vogt, J.E.: {Interpretability and Explainability: {A} Machine
  Learning Zoo Mini-tour}. Computing Research Repository ({CoRR})
  \textbf{abs/2012.01805} (2020)

\bibitem{Odena_artifacts}
Odena, A., Dumoulin, V., Olah, C.: {Deconvolution and Checkerboard Artifacts}.
  Distill  (2016)

\bibitem{Petsiuk_DRISE}
Petsiuk, V., Jain, R., Manjunatha, V., Morariu, V.I., Mehra, A., Ordonez, V.,
  Saenko, K.: {Black-box Explanation of Object Detectors via Saliency Maps}.
  In: Proceedings of the IEEE/CVF Conference on Computer Vision and Pattern
  Recognition ({CVPR}). pp. 11443--11452 (2021)

\bibitem{Ren_FasterRCNN}
Ren, S., He, K., Girshick, R.B., Sun, J.: {Faster {R-CNN:} Towards Real-Time
  Object Detection with Region Proposal Networks}. {IEEE} Transactions on
  Pattern Analysis Machine Intelligence ({PAMI})  \textbf{39}(6),  1137--1149
  (2017)

\bibitem{Rosenfeld_Elephant}
Rosenfeld, A., Zemel, R.S., Tsotsos, J.K.: {The Elephant in the Room}.
  Computing Research Repository ({CoRR})  \textbf{abs/1808.03305} (2018)

\bibitem{Samek_CVPRW21}
Samek, W., Montavon, G., Lapuschkin, S., Anders, C.J., M{\"{u}}ller, K.:
  {Explaining Deep Neural Networks and Beyond: {A} Review of Methods and
  Applications}. Proceedings of the {IEEE}  \textbf{109}(3),  247--278 (2021)

\bibitem{Selvaraju_GradCAM}
Selvaraju, R.R., Cogswell, M., Das, A., Vedantam, R., Parikh, D., Batra, D.:
  {Grad-CAM: Visual Explanations from Deep Networks via Gradient-Based
  Localization}. International Journal of Computer Vision  \textbf{128}(2),
  336--359 (2020)

\bibitem{Simonyan_Gradients}
Simonyan, K., Vedaldi, A., Zisserman, A.: {Deep Inside Convolutional Networks:
  Visualising Image Classification Models and Saliency Maps}. In: Bengio, Y.,
  LeCun, Y. (eds.) 2nd International Conference on Learning Representations
  ({ICLR}) Workshop Track Proceedings (2014)

\bibitem{Springenberg_GuidedBackpropagation}
Springenberg, J.T., Dosovitskiy, A., Brox, T., Riedmiller, M.A.: {Striving for
  Simplicity: The All Convolutional Net}. In: Bengio, Y., LeCun, Y. (eds.) 3rd
  International Conference on Learning Representations ({ICLR}) Workshop Track
  Proceedings (2015)

\bibitem{Sugawara_artifacts}
Sugawara, Y., Shiota, S., Kiya, H.: {Checkerboard artifacts free convolutional
  neural networks}. APSIPA Transactions on Signal and Information Processing
  \textbf{8}, ~e9 (2019)

\bibitem{Sundararajan_IG}
Sundararajan, M., Taly, A., Yan, Q.: {Axiomatic Attribution for Deep Networks}.
  In: Precup, D., Teh, Y.W. (eds.) Proceedings of the 34th International
  Conference on Machine Learning ({ICML}) 2017. Proceedings of Machine Learning
  Research, vol.~70, pp. 3319--3328. {Proceedings of Machine Learning Research
  (PMLR)} (2017)

\bibitem{Tan_EfficientDet}
Tan, M., Pang, R., Le, Q.V.: {EfficientDet: Scalable and Efficient Object
  Detection}. In: 2020 {IEEE/CVF} Conference on Computer Vision and Pattern
  Recognition ({CVPR}). pp. 10778--10787. Institute of Electrical and
  Electronics Engineers ({IEEE}) (2020)

\bibitem{Tomsett_sanity_checks_for_saliency_metrics}
Tomsett, R., Harborne, D., Chakraborty, S., Gurram, P., Preece, A.D.: Sanity
  checks for saliency metrics. In: Proceedings of the AAAI conference on
  artificial intelligence. pp. 6021--6029. {AAAI} Press (2020)

\bibitem{Toner_aisafety}
Toner, H., Acharya, A.: Exploring clusters of research in three areas of ai
  safety. Center for Security and Emerging Technology  (2022)

\bibitem{Tsunakawa_SSD_CRP}
Tsunakawa, H., Kameya, Y., Lee, H., Shinya, Y., Mitsumoto, N.: {Contrastive
  Relevance Propagation for Interpreting Predictions by a Single-Shot Object
  Detector}. In: 2019 International Joint Conference on Neural Networks
  ({IJCNN}). pp.~1--9. Institute of Electrical and Electronics Engineers
  ({IEEE}) (2019)

\bibitem{Matias_MarineDebris}
Valdenegro{-}Toro, M.: {Deep Neural Networks for Marine Debris Detection in
  Sonar Images}. Computing Research Repository ({CoRR})
  \textbf{abs/1905.05241} (2019)

\bibitem{Matias_MarineDebris_Dataset}
Valdenegro{-}Toro, M.: {Forward-Looking Sonar Marine Debris Datasets}. GitHub
  (2019), (Online accessed on 01 December 2021)

\bibitem{Yona_revisiting_sanity_checks}
Yona, G., Greenfeld, D.: Revisiting sanity checks for saliency maps. CoRR
  \textbf{abs/2110.14297} (2021)

\bibitem{Zeiler_DeconvNet}
Zeiler, M.D., Fergus, R.: {Visualizing and Understanding Convolutional
  Networks}. In: Fleet, D.J., Pajdla, T., Schiele, B., Tuytelaars, T. (eds.)
  Computer Vision -- {ECCV} 2014. Lecture Notes in Computer Science ({LNCS}),
  vol.~8689, pp. 818--833. Springer (2014)

\end{thebibliography}

\appendix
\section{Additional Sanity Check Results}

\begin{figure*}[!hb]
	\centering
	\includegraphics[width=\linewidth]{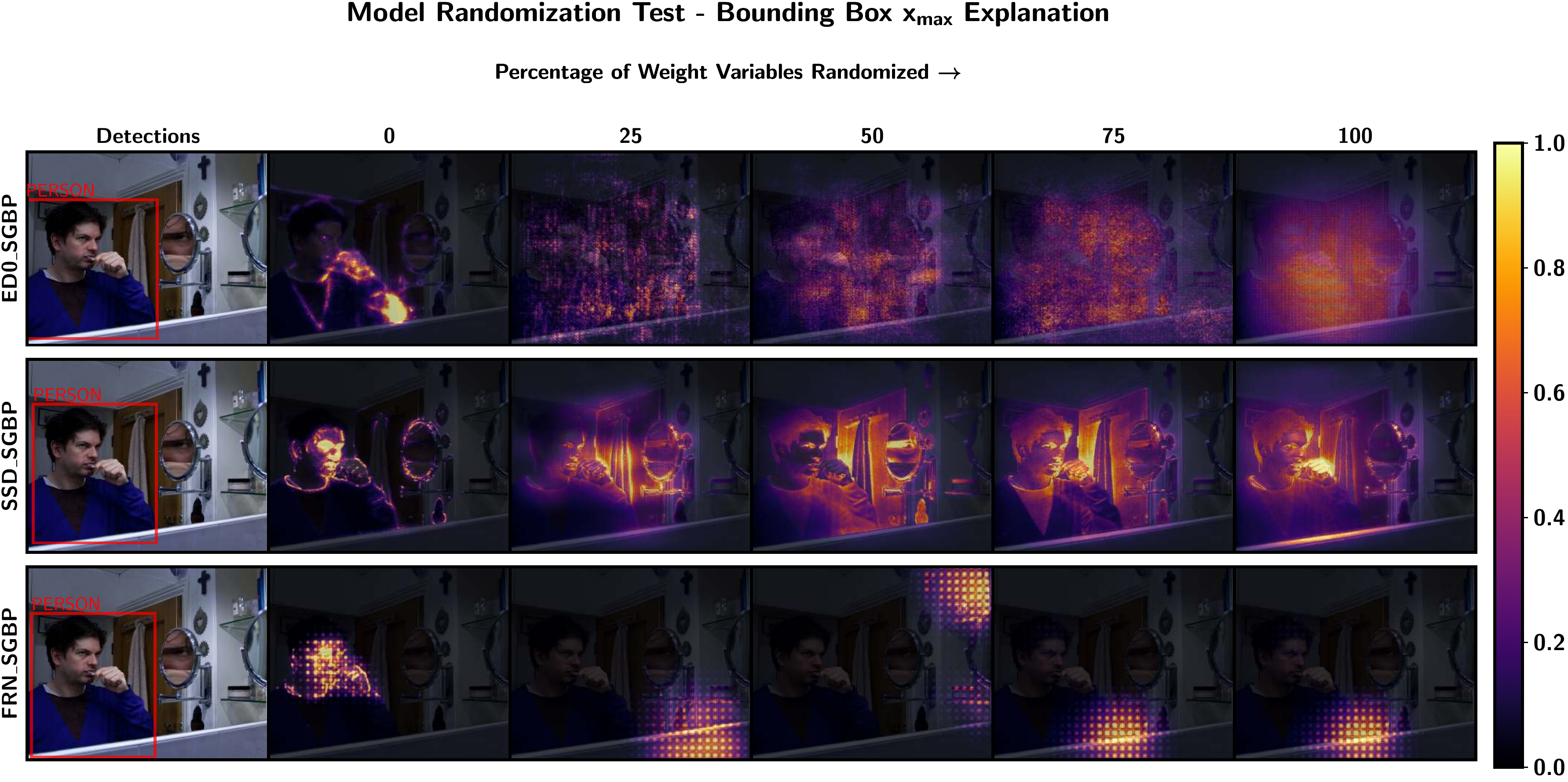}
	\caption{\label{fig:sanity_xmax_sgbp} Model randomization test for $x_{max}$ explanations (red-colored box) across different models using SGBP. The first column is the detection of interest that is explained in the consecutive columns. 
		The second column is the saliency map generated using the trained model without randomizing any parameters. 
		The second column highlights the important parts such as hands, eyes, and face. 
		The last column is the saliency map generated using a model with all parameters randomized. Note how FRN fails the randomization test.
	}
\end{figure*}

\begin{figure*}[!htb]
	\centering
	\includegraphics[width=\linewidth]{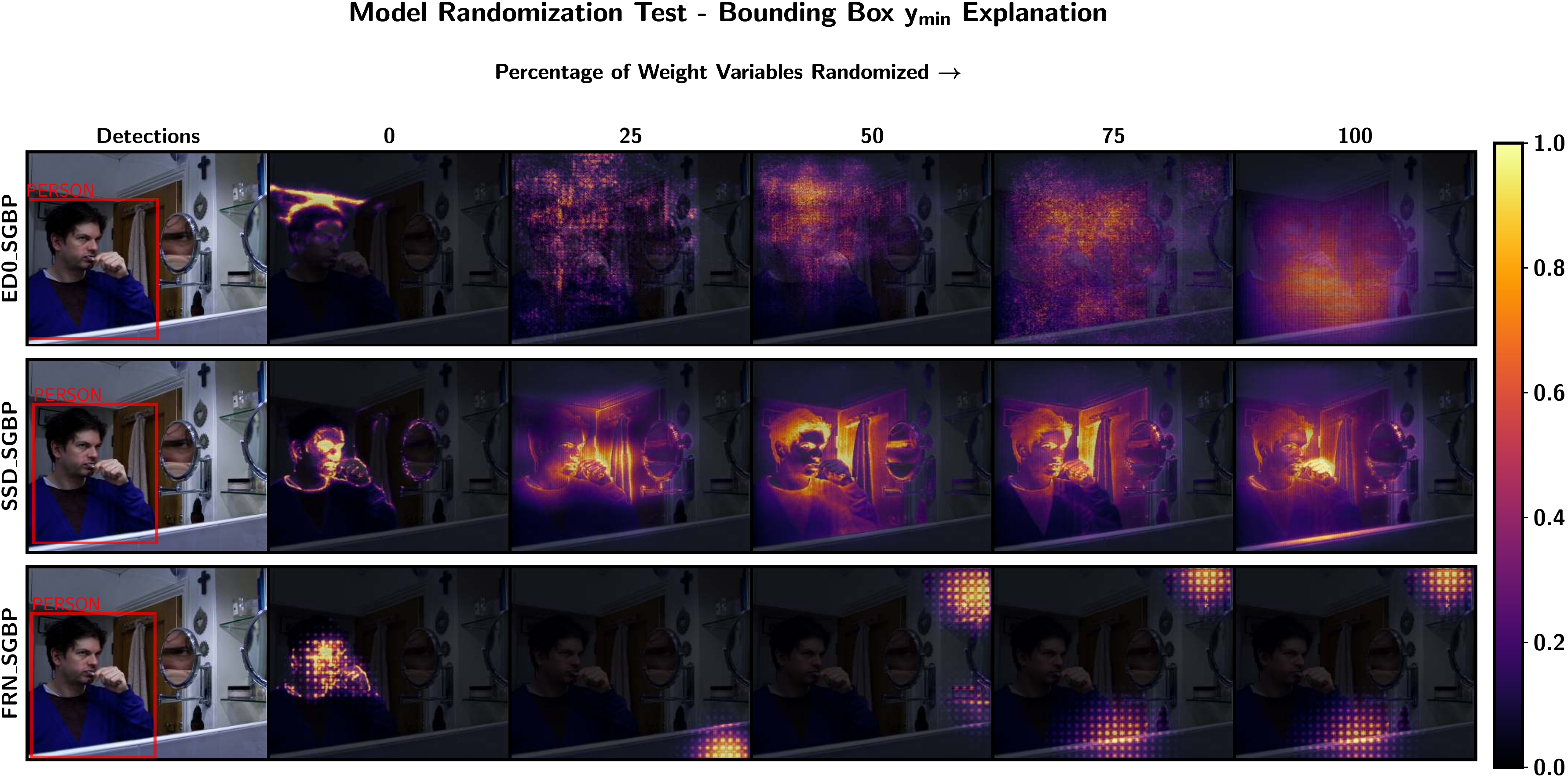}
	\caption{\label{fig:sanity_ymin_sgbp} Model randomization test for $y_{min}$ explanations (red-colored box) across different models using SGBP. The first column is the detection of interest that is explained in the consecutive columns. 
		The second column is the saliency map generated using the trained model without randomizing any parameters. 
		The second column highlights the important parts such as hands, eyes, and face. 
		The last column is the saliency map generated using a model with all parameters randomized. Note how FRN fails the randomization test.
	}
\end{figure*}

\begin{figure*}[!htb]
	\centering
	\mbox{} \hfill
	\includegraphics[width=\linewidth]{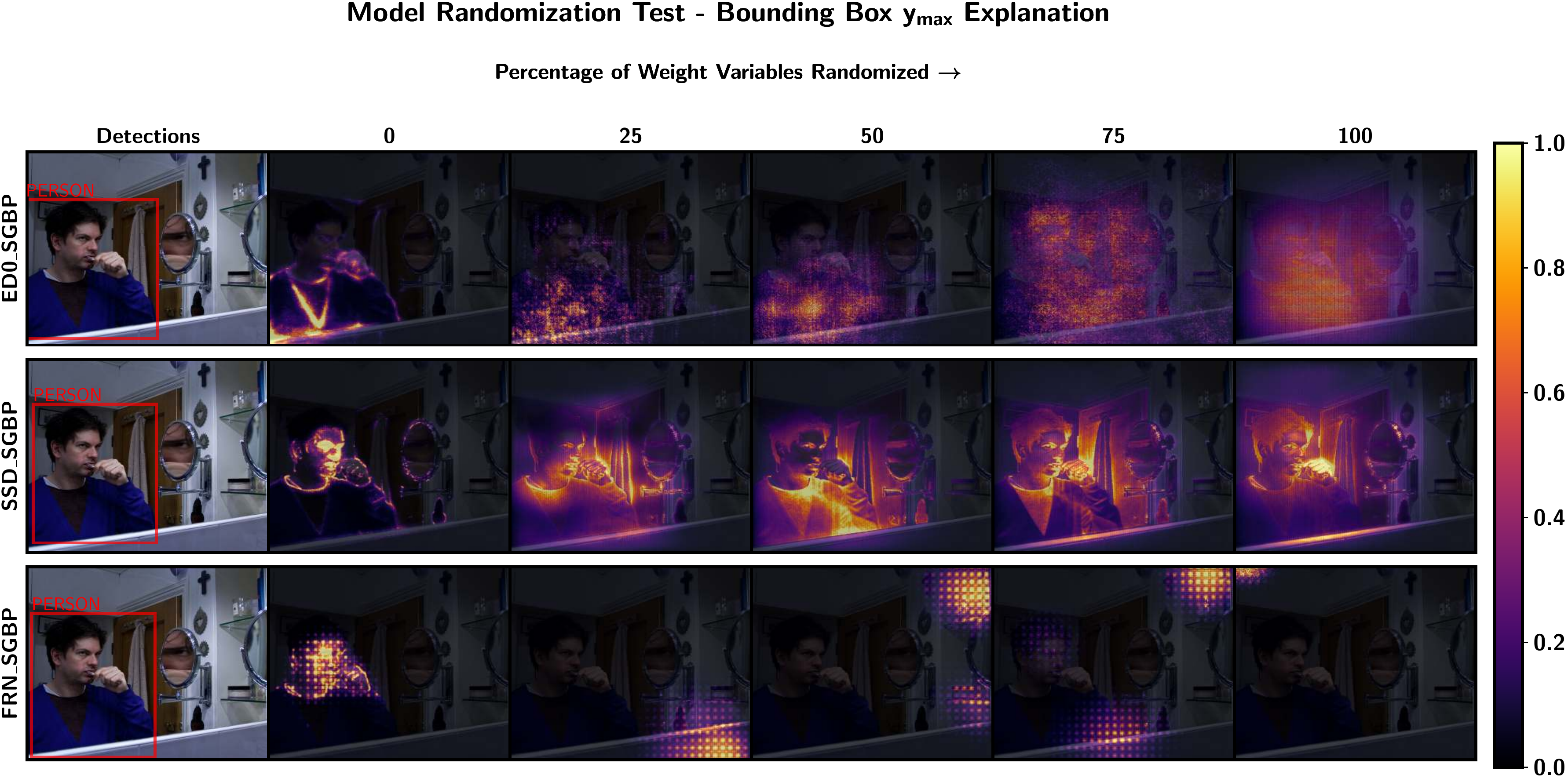}
	\caption{\label{fig:sanity_ymax_sgbp} Model randomization test for $y_{max}$ explanations (red-colored box) across different models using SGBP. The first column is the detection of interest that is explained in the consecutive columns. 
		The second column is the saliency map generated using the trained model without randomizing any parameters. 
		The second column highlights the important parts such as hands, eyes, and face. 
		The last column is the saliency map generated using a model with all parameters randomized. Note how FRN fails the randomization test.
	}
\end{figure*}

\end{document}